\pdfoutput=1

\documentclass[11pt, table]{article}

\usepackage[final]{acl}
\usepackage{booktabs}
\usepackage{times}
\usepackage{latexsym}
\usepackage{multirow}
\usepackage{rotating}
\usepackage{url}

\usepackage[T1]{fontenc}

\usepackage[utf8]{inputenc}

\usepackage{microtype}

\usepackage{inconsolata}

\usepackage{graphicx}
\usepackage{amsmath}
\usepackage{array}
\usepackage{amsfonts}
\usepackage{xcolor}
\usepackage{pgf}
\usepackage{collcell}
\usepackage{float}
\usepackage{subcaption}
\usepackage{enumitem}
\usepackage{xspace}
\usepackage{soul}
\usepackage{xcolor}
\sethlcolor{yellow!40}
\soulregister\ref7
\soulregister\citep7
\soulregister\citet7
\soulregister\cite7
\soulregister\bench0
\soulregister\S7

\newcommand{\bench}{\textsc{GUI-Primitives}\xspace}

\newcommand{\huggingface}{\raisebox{-1.5pt}{\includegraphics[height=1.05em]{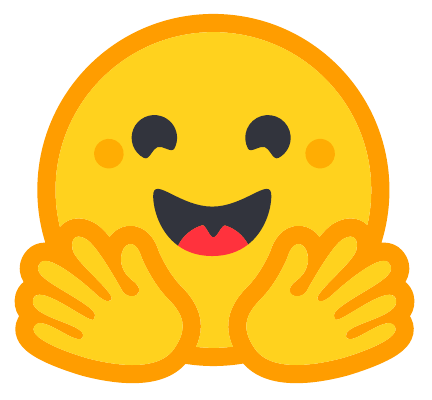}}\xspace}
\newcommand{\internet}{\raisebox{-1.5pt}{\includegraphics[height=1.05em]{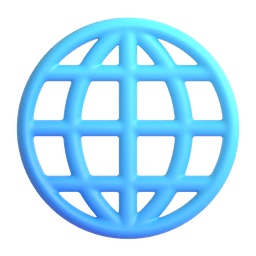}}\xspace}

\newcommand{\github}{\raisebox{-1.5pt}{\includegraphics[height=1.05em]{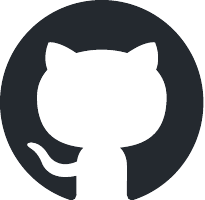}}\xspace}

\definecolor{high}{HTML}{ec462e}
\definecolor{low}{HTML}{ffffff}
\newcommand*{\opacity}{90}
\newcommand*{\MinNumber}{0.0}
\newcommand*{\MidNumber}{7.0}
\newcommand*{\MaxNumber}{15.0}

\newcommand{\colorcell}[1]{
    \pgfmathsetmacro{\value}{#1}

    \pgfmathparse{\value > \MaxNumber ? 1 : 0}
    \ifnum\pgfmathresult=1
        \hspace{-0.33em}\cellcolor{high!\opacity}{#1}
    \else
        \pgfmathparse{\value > \MidNumber ? 1 : 0}
        \ifnum\pgfmathresult=1
            \pgfmathparse{int(round(100*(\value - \MidNumber)/(\MaxNumber - \MidNumber)))}
            \xdef\tempa{\pgfmathresult}
            \hspace{-0.33em}\cellcolor{high!\tempa!yellow!\opacity}{#1}
        \else
            \pgfmathparse{int(round(100*(\MidNumber - \value)/(\MidNumber - \MinNumber)))}
            \xdef\tempa{\pgfmathresult}
            \hspace{-0.33em}\cellcolor{low!\tempa!yellow!\opacity}{#1}
        \fi
    \fi
}

\definecolor{darkblue}{rgb}{0, 0, 0.5}
\hypersetup{colorlinks=true, citecolor=darkblue, linkcolor=darkblue, urlcolor=darkblue}

\title{\bench{}: Diagnosing Spatial Reasoning Failures in Vision-Language GUI Grounding}

\newcommand{\aspace}{\hspace{0.7em}}
\newcommand{\usc}{$^{\heartsuit}$}
\newcommand{\isi}{$^{\spadesuit}$}

\newcommand{\qcri}{$^{\clubsuit}$}
\author{
    Md Abrar Jahin\usc \isi$^\dagger$ \aspace
    Md Rizwan Parvez\qcri$^\dagger$ \\
    \usc{}University of Southern California \aspace
    \isi{}USC Information Sciences Institute \\
    \qcri{}Qatar Computing Research Institute (QCRI)\\
    \texttt{jahin@usc.edu} \quad \quad \texttt{mparvez@hbku.edu.qa}\\
}

\begin{document}

\maketitle
\def\thefootnote{$\dagger$}\footnotetext{Corresponding Author(s)}
\def\thefootnote{$1$}

\begin{abstract}
\noindent
Computer-use agents ground natural-language instructions in screenshots to locate interface elements, yet existing benchmarks do not isolate whether models bind relational language to the correct element. We introduce \bench{}, a 994-item benchmark of contrastive instruction pairs over seven spatial relations in graphical user interfaces (left/right, above/below, containment, alignment, proximity, list ordinal, occlusion). Each pair holds the screenshot and anchor fixed while changing the relation expression, so the correct target moves between two designated candidates. Five annotators validate a 196-item subset (\mbox{$\kappa = 0.94$} well-formedness; \mbox{$\kappa = 0.79$} target selection). Nineteen vision-language models reach at most 32\% strict point-in-box accuracy. Because models emit unconstrained coordinates, we classify each prediction by the candidate region it falls within. Predictions fall outside both candidates on 60--92\% of items. Conditional on falling within a candidate region, target selection reaches 0.82--0.90 for horizontal position, vertical position, proximity, and list ordinal, but does not differ significantly from 0.50 for containment and occlusion: most failures reflect candidate localization rather than relation understanding. Across ten models, benchmark accuracy correlates with ScreenSpot-Pro accuracy (Spearman \mbox{$\rho = +0.74$}), an exploratory association at this sample size. Marking the two designated candidates raises selection accuracy by 35--57 percentage points, an oracle diagnostic that supplies the candidate set rather than a deployable method. We release the benchmark, predictions, and code.
\end{abstract}

\renewcommand{\arraystretch}{1.2}
\begin{table}[!b]
    \small
    \setlength{\tabcolsep}{2pt}
    \resizebox{\linewidth}{!}{
    \begin{tabular}{cll}
         \github & \textbf{Source Code:} & \href{https://github.com/Abrar2652/gui-primitives/}{\path{https://github.com/Abrar2652/gui-primitives/}} \\
         \huggingface & \textbf{Dataset:} & \href{https://huggingface.co/datasets/kagnlp/gui-primitives/}{\path{https://huggingface.co/datasets/kagnlp/gui-primitives/}} \\
         \internet & \textbf{Website:} & 
         \href{https://abrar2652.github.io/gui-primitives/}{https://abrar2652.github.io/gui-primitives/}
    \end{tabular}
    }
\end{table}
\renewcommand{\arraystretch}{1}

\section{Introduction}
\label{sec:intro}
A new class of \emph{computer-use agents} now reads a screenshot, decides what to do, and emits a click coordinate or a keystroke \citep{xie2024osworld,cheng2024seeclick,qin2025uitars,gou2025uground,wu2025guiactor, wang2025opencua}. These agents are beginning to be deployed; their first job, before any agent loop, is the same job a person does when they open a window: read the screen, find the right control, and put the pointer in the right place. We call this \emph{GUI grounding}, and a growing body of literature shows that frontier vision-language models still trail humans in this task \citep{li2025screenspotpro, yang2025gta1}.

A natural question follows. If we strip the agent away and ask a present-day VLM only to handle the elementary spatial reasoning that GUI grounding requires: to the right of \texttt{Save}, inside the \texttt{Layers} panel, the third item in the menu, does it succeed? Existing benchmarks do not isolate this skill. End-to-end benchmarks like OSWorld \citep{xie2024osworld} combine grounding with planning and action; coarse grounding benchmarks like ScreenSpot \citep{cheng2024seeclick} and ScreenSpot-Pro \citep{li2025screenspotpro} measure whether the click lands within the target box but do not separate failures by spatial primitive. As a result, when an agent clicks the wrong button, we do not yet know \emph{which} elementary
skill broke.

We close this gap with a controlled diagnostic. \bench{} is a benchmark of 994 contrastive-pair questions over seven spatial relations in GUI screenshots. Each question is paired with a twin that keeps the screenshot and anchor fixed and changes the relation expression (\texttt{left} $\leftrightarrow$ \texttt{right}, \texttt{inside} $\leftrightarrow$ \texttt{outside}, and so on), so that the correct target moves to the other designated candidate. In 253 of 497 pairs, the surrounding instruction template also varies lexically (verb, head noun, anchor quoting); this variation is not confounded with the relation term (Appendix~\ref{app:sources}). This construction, adapted from the minimal-pair tradition in linguistic evaluation \citep{linzen2016agreement, marvin2018minimalpair, kamath2023whatsup}, controls for screenshot-specific salience and fixed answer preferences: a model that always clicks the most salient or central element scores correctly on one twin and incorrectly on the other. We evaluate 19 vision-language models spanning open-weight and proprietary systems. Five annotators independently validated a 196-item subset (Fleiss \mbox{$\kappa = 0.94$} for well-formedness) and selected the instruction-consistent candidate on 96.9\% of retained items; this is annotator target-selection accuracy on a two-candidate task, not a model-matched grounding score.

The evaluation yields three main findings. Claude Opus 4.7, the strongest model, reaches 31\% strict point-in-box accuracy on the human-clean subset. On four of seven primitives (containment, occlusion, alignment, proximity), every model we test falls below the two-candidate reference level of 0.50. Candidate-level analysis (\S\ref{subsec:perprim}) indicates that this below-reference accuracy is primarily attributable to predictions outside both candidate regions rather than to the selection of the contrastive element. On real desktop screenshots from UI-Vision \citep{uivision}, accuracy is lower still, and candidate-level analysis attributes this primarily to predictions outside both candidate regions (\mbox{$96.2\%$} of real-subset predictions) rather than to relation errors. Pair consistency is below the value expected under independent responses because carrier-template variation and off-candidate prediction also contribute to this dependence; therefore, candidate-level analysis (\S\ref{subsec:perprim}) is required to interpret it.

Yet the diagnosis is not only a problem statement. Three results follow. First, the benchmark tracks downstream performance. Across ten models for which we also run ScreenSpot-Pro, \bench{} accuracy correlates with grounding accuracy at Spearman $\rho = +0.74$ ($p = 0.015$). The correlation is stable across the real-screenshot and synthetic slices of \bench{}. Item-level logistic regression on $n = 15{,}810$ model--item pairs corroborates this association (pseudo-$R^{2} = 0.40$); we report it conservatively. Second, we test three training-free interventions: Set-of-Mark (SoM) prompting \citep{yang2023setofmark}, primitive-aware chain-of-thought (CoT), and activation steering \citep{kang2025localization,sivakumar2025steervlm}. Marking the two designated candidates (an oracle two-candidate diagnostic) raises selection accuracy by 35--57 percentage points, lifting GPT-5 from 30\% to 87\% and OS-Atlas from 10\% to 52\%; CoT and activation steering yield no significant gain. Because the intervention supplies the candidate set that the baseline task requires the model to locate, it is an upper bound on candidate discrimination rather than a deployable method (\S\ref{sec:discussion}). Third, three shortcut controls (blank canvas, shuffled instructions, heavy blur) reduce accuracy as expected for a grounded benchmark; the blur control produces a significant but sub-threshold drop, and the negative results for CoT and activation steering are reported alongside the positive result for SoM.

We make four contributions. First, we introduce \bench{}, a 994-item contrastive-pair diagnostic benchmark covering seven spatial relations in GUI screenshots, together with a five-annotator verified 196-item core (\(\kappa = 0.94\)). Second, we evaluate 19 models across five families and localize the dominant failure to candidate localization, predictions fall outside both candidate regions on 60--92\% of items, with relation-word errors confined to containment and occlusion, alongside a 65-point gap from human performance. Third, we show that primitive competence correlates with downstream GUI grounding on ScreenSpot-Pro (Spearman \(\rho = +0.74\), \(p = 0.015\)). Fourth, we find that SoM transfers across both open and closed models, improving GPT-5 by 57 points, whereas CoT prompting and activation steering do not; we release \bench{}, all model predictions, and analysis code.


\begin{figure*}[!t]
  \centering
  \includegraphics[width=0.8\linewidth]{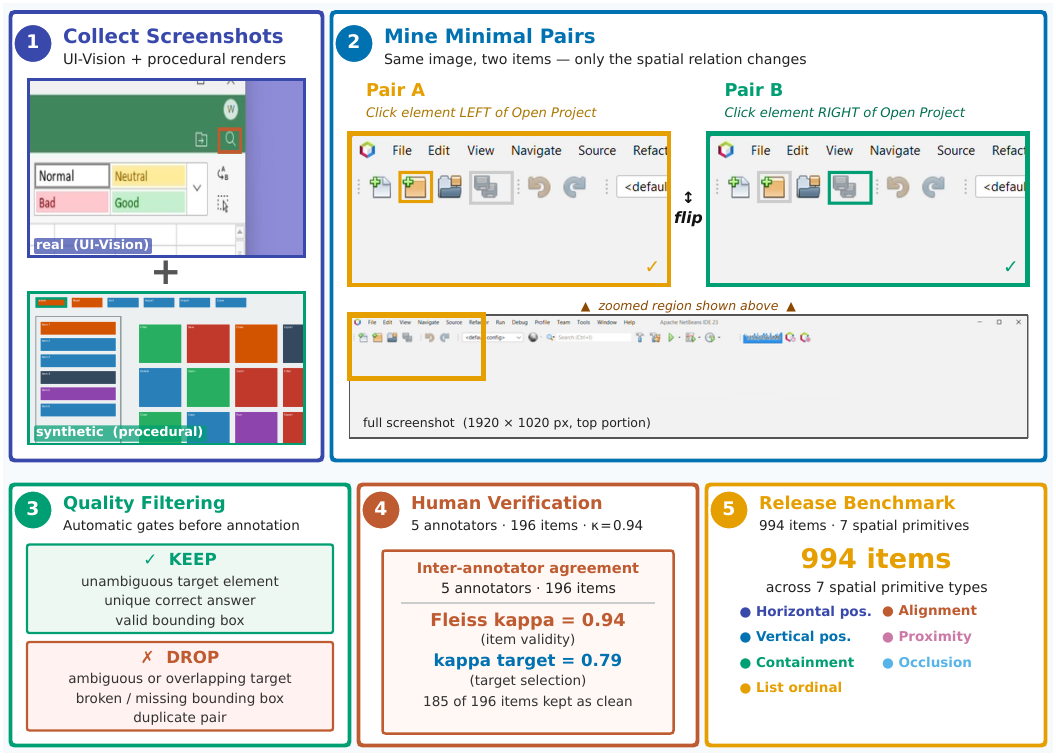}
  \caption{\textbf{\bench{} dataset construction pipeline.} Two source corpora: real desktop screenshots from UI-Vision and controlled synthetic stimuli from our procedural renderer, feed into per-primitive minimal-pair miners, pass through automatic quality gates, are cross-checked by a five-annotator human verification pass (Fleiss $\kappa{=}0.94$), and released as the 994-item benchmark. Stage colors follow the primitive palette used throughout the paper.}
  \label{fig:construction-banner}
\end{figure*}

\section{Related Work}
\label{sec:related}

\paragraph{GUI grounding and computer-use agents.}
A line of recent work targets agents that operate real software. OSWorld \citep{xie2024osworld} provides a real Linux and Windows VM environment with execution-based scoring on 369 tasks; a subsequent human-verified release closes ambiguous tasks \citep{osworldverified2025}. ScreenSpot \citep{cheng2024seeclick} and ScreenSpot-Pro \citep{li2025screenspotpro} measure pixel-grounding in isolation, the latter targeting professional software at high resolution; MMBench-GUI \citep{wang2025mmbenchgui} adds a hierarchical multi-platform evaluation. Grounding models trained on screenshots include SeeClick, OS-Atlas \citep{wu2024osatlas}, UI-TARS \citep{qin2025uitars}, UGround \citep{gou2025uground}, Aria-UI \citep{yang2025ariaui}, GUI-Actor \citep{wu2025guiactor}, Jedi \citep{xie2025jedi}, OpenCUA \citep{wang2025opencua}, and the test-time scaling agent GTA1 \citep{yang2025gta1}. Closest in spirit, GUI-Cursor \citep{zhao2025guicursor} probes a cursor-based agent's spatial reasoning under reinforcement learning, and Beyond Semantics \citep{qi2025beyondsemantics} argues that large vision-token norms suppress positional encoding, a mechanistic candidate for our failures. These works ask whether models land the click or how to train them; we ask which elementary spatial skill they lack when they miss.

\paragraph{Spatial reasoning in vision-language models.}
Several benchmarks reveal that VLMs struggle with spatial relations even on natural images. \textsc{What's-Up} \citep{kamath2023whatsup} uses controlled minimal pairs to show that contrastive VLMs handle left/right and on/under poorly despite high captioning accuracy. VSR \citep{liu2023vsr} catalogs 66 spatial relations in natural images; BLINK \citep{fu2024blink} bundles 14 perceptual primitives and finds even GPT-4V well below humans; CV-Bench \citep{tong2024cambrian} shows that 2D/3D spatial primitives are bottlenecked by the visual encoder. SpatialVLM \citep{chen2024spatialvlm} addresses these gaps by co-training on synthetic 3D-grounded data. RocketScience \citep{hoehing2025rocketscience} uses contrastive pairs to disentangle object localization from spatial reasoning, and finds that only top reasoning models succeed. \bench{} extends this controlled methodology to GUI screenshots, where the visual prior is starkly different: text labels everywhere, near-identical icons, dense layout.

\paragraph{Minimal-pair evaluation.}
Minimal pairs originate in psycholinguistic evaluation \citep{linzen2016agreement,marvin2018minimalpair}, where a single lexical or syntactic edit creates a contrast whose disambiguation requires understanding rather than surface statistics. \textsc{What's-Up}
brought the design to vision-language. We adopt the same logic for GUI screenshots and add pair-consistency \citep{kamath2023whatsup} as an internal diagnostic of relational failure (\S\ref{subsec:pair-consistency}).

\paragraph{Inference-time interventions.}
SoM prompting \citep{yang2023setofmark} overlays numbered marks on visual regions so the model names the mark rather than regressing a coordinate. SoM-LLaVA \citep{yan2024somllava} shows that open-source MLLMs need explicit SoM training data to match closed models. Visual prompting is now a standard tool in GUI-agent stacks, with recent training-free zoom and modality-aware variants: DiMo-GUI \citep{wu2025dimogui} decouples text and icon modalities and applies test-time scaling; RegionFocus \citep{luo2025regionfocus} reaches 61.6\% on ScreenSpot-Pro with a Qwen2.5-VL-72B backbone by iteratively zooming into a focal region. Activation and attention steering, including the localization-head method of \citet{kang2025localization} and the SteerVLM framework \citep{sivakumar2025steervlm}, modify model internals at inference time without weight updates. We test one representative from each family.

\section{The \bench{} Benchmark}
\label{sec:bench}
\bench{} is designed for one purpose: to ask whether a vision-language model has each of the seven elementary spatial skills that GUI grounding requires. Three design choices follow from this purpose: the seven primitives (\S\ref{subsec:primitives}), the minimal-pair construction (\S\ref{subsec:minpair}), and the human verification protocol (\S\ref{subsec:verification}). The full construction pipeline is summarized in Figure~\ref{fig:construction-banner}.

\subsection{Seven elementary primitives}
\label{subsec:primitives}
We chose primitives that are unambiguous, common in everyday GUI language, and decomposable. Table~\ref{tab:primitives} lists all seven together with their relation words and chance level. The primitives split into three groups. \emph{Cardinal direction} covers horizontal and vertical relative position. \emph{Topology} covers containment (inside or outside a panel). \emph{Layout} covers alignment (sharing a row or a column), proximity (which of several elements is nearest), list ordinal (the third item of a menu), and occlusion (visible or partly covered by a pop-up).

\paragraph{Why these seven?}
We require each primitive to be lexically realizable in a short user instruction, separable by a single relation-expression change into a contrastive pair, and a basic building block of compositional GUI relations. The seven jointly cover the four broad classes cataloged by prior natural-image work: cardinal direction, topology, ordinality, and layout-distance \citep{liu2023vsr,kamath2023whatsup,hoehing2025rocketscience}. We do not claim coverage of every spatial
relation, only of the elementary ones from which the rest compose.

\begin{table}[ht]
\small
\centering
\caption{The seven elementary spatial primitives in \bench{}, with their canonical relation words and two-candidate reference levels. All seven primitives, including \emph{list-ordinal}, are realized as binary minimal-pair contrasts (71 items per side), so the reference level is $0.50$ throughout; \S\ref{sec:protocol} explains why this is a forced-choice reference rather than the chance level of unconstrained clicking.}
\label{tab:primitives}
\setlength{\tabcolsep}{3pt}
\renewcommand{\arraystretch}{1.1}
\begin{tabular}{@{}lllc@{}}
\toprule
\textbf{Primitive} & \textbf{Relation words} & \textbf{Group} & \textbf{Ref.\ level} \\
\midrule
rel-pos-horizontal & left / right        & cardinal & 0.50 \\
rel-pos-vertical   & above / below       & cardinal & 0.50 \\
containment        & inside / outside    & topology & 0.50 \\
alignment          & same row / col.\    & layout   & 0.50 \\
proximity          & nearest / farthest  & layout   & 0.50 \\
list-ordinal       & first / second      & layout   & 0.50 \\
occlusion          & visible / hidden    & layout   & 0.50 \\
\bottomrule
\end{tabular}
\end{table}

\subsection{Minimal-pair construction}
\label{subsec:minpair}
Every item in \bench{} appears as one half of a contrastive pair: the two members share one screenshot and one anchor element, and the relation expression changes, with the correct target as the \emph{contrastive} element (Figure~\ref{fig:teaser}, details in Appendix~\ref{app:minpair}). This construction removes the shortcuts that inflate scores on coarser benchmarks: a model that always selects the most central, salient, or frequently labeled element earns one point and loses one on a pair, so the pair-level score is zero. We measure this directly as \emph{pair-consistency} (\S\ref{subsec:pair-consistency}). Three representative pairs for each of the remaining six primitives are shown in Appendix~\ref{app:more-teasers}.

\begin{figure}[ht]
  \centering
  \includegraphics[width=\columnwidth]{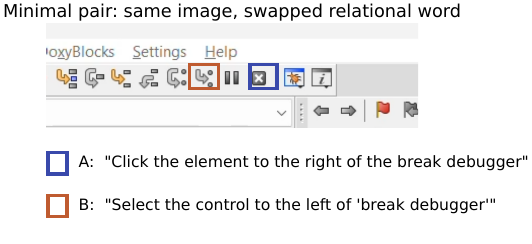}
  \caption{A contrastive pair from \bench{} (details in Appendix~\ref{app:more-teasers}). The screenshot is fixed; one word flips, and the correct target shifts to the other candidate. A model exploiting an answer prior passes one twin and fails the other.}
  \label{fig:teaser}
\end{figure}

\subsection{Sources: real and synthetic screenshots}
\label{subsec:sources}
\bench{} draws from two sources (Appendix~\ref{app:sources} and \ref{app:distribution}), both reported separately throughout. The \emph{real} arm ($n = 290$) uses element-grounding annotations from UI-Vision \citep{uivision} on Web, MacOS, VS Code, Office, and professional applications, covering five primitives whose annotations are recoverable from element labels: horizontal and vertical relative position, alignment, proximity, and a 24-item list-ordinal slice mined from real menus and lists. The \emph{synthetic} arm ($n = 704$) is procedurally rendered with known positions, controlled distractors, and a fixed layout language; it is the only way to obtain \emph{containment} and \emph{occlusion} cleanly, and it supplies the remaining \emph{list-ordinal} items, since real screenshot annotations do not ship the parent--child or overlay--target metadata. We treat the synthetic arm as a controlled-stimulus arm in the \textsc{What's-Up} tradition and disclose every result split by source.

\subsection{Human verification}
\label{sec:humanverif}
\label{subsec:verification}
Five annotators independently judged a 196-item stratified subset (\textit{human-verified core}), answering two questions per item: well-formedness and correct target (Appendix~\ref{app:verification}). Agreement is high: Fleiss $\kappa_{Q1} = 0.942$ (``almost perfect'' on the Landis--Koch scale, \citealp{landis1977measurement}) and $\kappa_{Q2} = 0.787$ (``substantial''). Eleven items judged invalid by the majority were dropped, leaving the 185-item \textit{human-clean core}; human accuracy on it is 96.9\%. Annotator instructions are in Appendix~\ref{app:human-annotation}.

\section{Models and Protocol}
\label{sec:protocol}
\paragraph{Models.} We evaluate 19 vision-language models: seven proprietary models accessed through commercial APIs (Claude Opus 4.7, Sonnet 4.6, Haiku 4.5, GPT-5, GPT-4.1, GPT-4o-mini, Gemini 3.1 Flash Lite); three Qwen-family open (Qwen2.5-VL-7B, Qwen2-VL-7B, OS-Atlas-Base-7B); InternVL3-8B; Meta Llama-3.2-11B-Vision; three sizes of Gemma~3; two PaliGemma~2 variants; and MiniCPM-V. Identifiers and endpoints are in Appendix~\ref{app:models}. Pixtral-12B and Idefics3 8B are excluded because their chat template formats are incompatible with our generic
wrapper.

\paragraph{Prompting and decoding.}
All models use greedy decoding ($T = 0$) and a fixed seed. Per-family coordinate conventions (pixel, normalized $[0,1]$, normalized $[0,1000]$) are handled by a robust parser whose unit tests are bundled with the artifact. The three closed providers
silently downsample large screenshots server-side (Anthropic $\le 1568$~px, OpenAI $\approx 1024$ on the long side, Gemini variable); we pre-resize to a known long side and rescale predicted coordinates back to the original pixel space. The rescale was
needed on a large fraction of items (e.g., 1{,}297 of 1{,}581 Sonnet ScreenSpot-Pro records); Appendix~\ref{app:rescale} documents the procedure and the per-model record counts.

\paragraph{Metrics.}
The primary metric, following ScreenSpot-Pro, is \emph{point-in-box} accuracy: a prediction $\hat{p} = (\hat{x}, \hat{y})$ is correct iff $\hat{x} \in [x_{1}, x_{2}]$ and $\hat{y} \in [y_{1}, y_{2}]$ in the original pixel frame. \emph{Loose} accuracy (prediction within $2\times$ target diagonal of the box center) separates sub-pixel imprecision from element misidentification on small icons. All accuracies carry bootstrap 95\% CIs (2{,}000 resamples); paired comparisons use McNemar \citep{mcnemar1947note} with Holm--Bonferroni \citep{holm1979simple} correction across the seven primitives. All seven primitives are binary minimal-pair contrasts (Table~\ref{tab:primitives}), so $0.50$ serves as a \emph{two-candidate reference level} throughout. Because models emit an unconstrained coordinate rather than selecting among candidates, $0.50$ is a forced-choice reference rather than the chance level of the prediction task, and a prediction can fall outside both candidate regions. We flag a cell as \emph{below the reference level} only when the upper bootstrap bound falls below $0.50$. To distinguish selection of the contrastive element from predictions outside both regions, we classify every baseline prediction as target, distractor, neither, or invalid in Appendix~\ref{app:candidate} (further details in Appendix~\ref{app:protocol}).

\section{Diagnostic Results}
\label{sec:results}
We report results in five parts, in this order: (i) controls that validate the benchmark, (ii) overall accuracy and the human gap, (iii) per-primitive structure of failure, (iv) the real-vs-synthetic
decomposition, and (v) pair-consistency as direct evidence of relational failure.

\subsection{Shortcut controls validate the benchmark}
\label{subsec:controls}
\begin{figure}[ht]
  \centering
  \includegraphics[width=0.8\columnwidth]{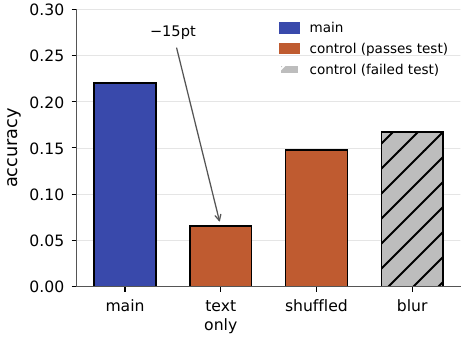}
  \caption{Shortcut controls (Qwen2.5-VL-7B, $n = 994$). Text-only and shuffled drops are both highly significant; blur produces a smaller but significant drop, below the pre-registered ten-point threshold.}
  \label{fig:controls}
\end{figure}

A diagnostic is only as good as the shortcuts it forbids (Appendix~\ref{app:controls}). Figure~\ref{fig:controls} reports three controls on Qwen2.5-VL-7B over the full 994-item benchmark.

Pairing the instruction with a \emph{blank canvas} drops accuracy from 22.0\% to 6.5\% ($\Delta = -15.5$~pt, $p = 2.8 \times 10^{-27}$); \emph{shuffling} the screenshot to one from a different item of the same primitive drops it to 14.8\% ($\Delta = -7.2$~pt, $p = 1.3 \times 10^{-11}$). Both behave as a grounded model should: when the visual or the matching instruction is removed, the task collapses. \emph{Heavy Gaussian blur} ($\sigma = 12$~px) drops accuracy only to 16.7\% ($\Delta = -5.3$~pt, $p = 8.5 \times 10^{-5}$) significant but below our pre-registered ten-point threshold. VLM grounding appears to rely more on global layout than on fine-grained OCR of text labels, since removing high-frequency detail removes only a modest fraction of the signal. However, blur alone cannot disentangle failures in OCR, spatial encoding, and cross-modal binding; targeted ablations are discussed in Appendix~\ref{app:binding}.

\subsection{Overall accuracy and the human gap}
\label{subsec:overall}

\begin{table*}[ht]
\centering
\small
\caption{Overall accuracy on \bench{} for 12 representative models, alongside accuracy on the human-clean 185-item subset. The strongest model (Claude Opus 4.7) reaches 32\% on the clean core, a 65-point gap below human accuracy (96.9\%). All 19 models are in Appendix~\ref{app:full-results}.}
\begin{tabular}{@{}lcccc@{}}
\toprule
\textbf{Model} & \textbf{Family} & \textbf{Full ($n{=}994$)} &
  \textbf{Core ($n{=}196$)} & \textbf{Clean ($n{=}185$)} \\
\midrule
Claude Opus 4.7              & Anthropic     & 0.313 & 0.311 & 0.324 \\
GPT-5 (reasoning)            & OpenAI        & 0.265 & 0.291 & 0.297 \\
Claude Sonnet 4.6            & Anthropic     & 0.222 & 0.260 & 0.270 \\
Claude Haiku 4.5             & Anthropic     & 0.239 & 0.250 & 0.249 \\
Qwen2.5-VL-7B                & Qwen2.5-VL    & 0.220 & 0.245 & 0.249 \\
Gemma 3-27B                  & Google open   & 0.134 & ---   & 0.157 \\
OS-Atlas-Base-7B             & Qwen2-VL/GUI  & 0.101 & 0.107 & 0.108 \\
GPT-4.1                      & OpenAI        & 0.096 & ---   & 0.108 \\
InternVL3-8B                 & OpenGVLab     & 0.095 & ---   & 0.097 \\
Gemini 3.1 Flash Lite        & Google closed & 0.073 & 0.077 & 0.076 \\
GPT-4o-mini                  & OpenAI        & 0.068 & 0.077 & 0.076 \\
Qwen2-VL-7B                  & Qwen2-VL      & 0.038 & 0.046 & 0.049 \\
Llama-3.2-11B-Vision         & Meta mllama   & 0.012 & 0.005 & 0.005 \\
\midrule
\textbf{Human}                & ---           & ---   & ---   & \textbf{0.969} \\
\bottomrule
\end{tabular}

\label{tab:overall}
\end{table*}

\begin{figure}[ht]
  \centering
  \includegraphics[width=0.9\columnwidth]{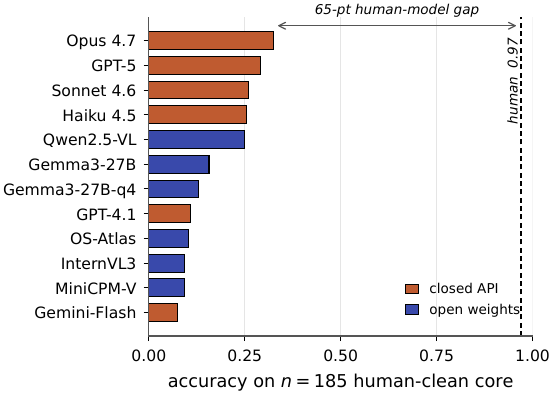}
  \caption{Human-clean core (185 items, 5 annotators,
    $\kappa = 0.94$). Human accuracy is 96.9\%. The strongest model
    is Claude Opus 4.7 at 32.4\%.}
  \label{fig:gap}
\end{figure}

Table~\ref{tab:overall} reports the headline numbers, and Figure~\ref{fig:gap} visualizes the gap from humans.

All evaluated models perform substantially below the annotator target-selection accuracy of 96.9\%, though the two protocols are not matched (\S\ref{sec:humanverif}): the strongest system (Claude Opus 4.7) reaches 31--32\% strict point-in-box accuracy and the 65-point gap holds on both the full benchmark and the cleaned 185-item core. Scale alone does not close it; the four strongest proprietary models cluster in a narrow 24--31\% range, and the 7B open Qwen2.5-VL-Instruct is statistically tied with closed Claude Haiku 4.5 on the human-verified core (24.5\% vs.\ 25.0\%) despite being roughly two orders of magnitude smaller and runnable on a single GPU. The rank order is stable across the full 994-item benchmark, the 196-item human-verified core, and an LLM-judge-verified 100-item subset (Appendix~\ref{app:full-results}), so the gap is not a noisy-item artifact.

\subsection{Per-primitive structure of failure}
\label{subsec:perprim}

\begin{figure}[ht]
  \centering
  \includegraphics[width=\columnwidth]{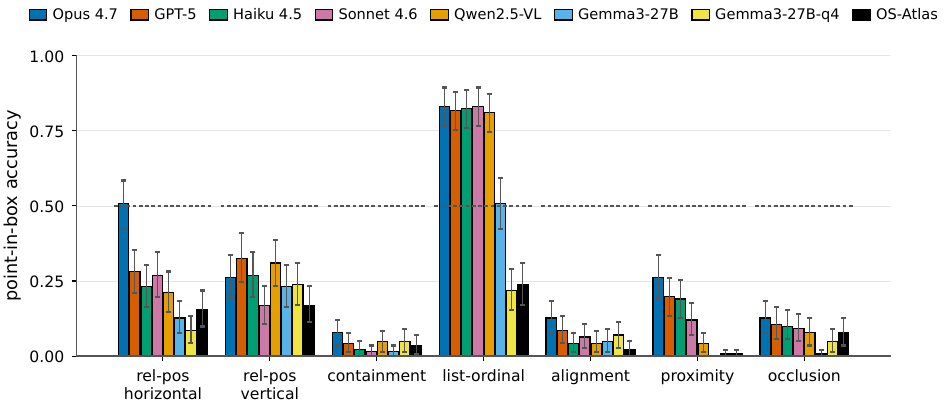}
  \caption{Per-primitive accuracy across eight representative models, with bootstrap 95\% confidence intervals. Dashed lines mark the two-candidate reference level (0.50 for all primitives; Table~\ref{tab:primitives}). On \texttt{containment}, \texttt{occlusion}, \texttt{alignment}, and \texttt{proximity}, every model falls below this reference, including the four strongest proprietary models; candidate-level analysis (Appendix~\ref{app:candidate}) attributes this below-reference accuracy primarily to predictions outside both candidate regions.}
  \label{fig:perprim}
\end{figure}

Figure~\ref{fig:perprim} unpacks the overall accuracy by primitive (Appendix~\ref{app:overall}). The picture is uneven and informative. One primitive, \texttt{list-ordinal}, is well solved by the top five models (Claude family, GPT-5, and Qwen2.5-VL), all achieving $0.80$--$0.83$ against the $0.50$ two-candidate reference (Table~\ref{tab:primitives}). The shared task here is counting list positions, which the language model can do once it identifies the list. The other six primitives are uniformly hard. On \texttt{rel-pos-horizontal}, Claude Opus 4.7 reaches $0.51$, at the reference level; every other model is below. On \texttt{rel-pos-vertical}, the best model is GPT-5 at $0.32$. On \texttt{containment}, \texttt{occlusion}, \texttt{alignment}, and \texttt{proximity}, \emph{the upper 95\% bootstrap bound for every model we test sits below the reference level} of $0.50$, including all four strongest proprietary models. Because $0.50$ is a two-candidate reference level rather than the chance level of unconstrained coordinate prediction, we classify all 18{,}886 baseline predictions by the candidate region they fall in (Appendix~\ref{app:candidate}). Predictions fall outside both candidate regions on 60--92\% of items, and among predictions inside a candidate region, the target is selected at least as often as the distractor on every primitive. For \texttt{containment} and \texttt{occlusion} the within-region selection rate is statistically indistinguishable from $0.50$ (0.548 and 0.554), indicating no measurable relation-word signal on these two primitives. For \texttt{rel-pos-horizontal}, \texttt{rel-pos-vertical}, and \texttt{proximity} the target is selected on 0.89--0.90 of within-region predictions, indicating that the deficit is candidate localization rather than relational interpretation; the \texttt{alignment} estimate varies with the classification rule (\S\ref{app:candidate}) and we do not interpret it. The associated directional prior is quantified in Appendix~\ref{app:sensitivity}, and its role is discussed in \S\ref{sec:discussion}. Model-specific coordinate-frame anomalies are also quantified in Appendix~\ref{app:sensitivity}.

\subsection{Real versus synthetic}
\label{subsec:realsyn}
We next separate two candidate explanations for the per-primitive results in \S\ref{subsec:perprim}: small real-GUI target size and the relative proportion of synthetic items. The full strict/loose split by source is in Appendix~\ref{app:realsyn-table} and \ref{app:realsyn}. On synthetic screenshots, \emph{loose} accuracy is $0.60$--$0.76$ even when strict accuracy is much lower, indicating that synthetic errors are largely sub-pixel imprecision rather than element misidentification. On UI-Vision real screenshots, strict accuracy is essentially zero for every model on every primitive, and \emph{loose} accuracy also stays low ($6$--$23\%$). Candidate-level analysis (Appendix~\ref{app:candidate}) shows that 96\% of real-screenshot predictions fall outside both the target and distractor regions, while among predictions inside a candidate region, the target is selected on 0.75 pooled (0.91 for the two strongest models). Real-screenshot accuracy is therefore primarily attributable to candidate localization, with target size a contributing factor (Appendix~\ref{app:sensitivity}), rather than to inverted relational interpretation.

\begin{figure}[ht]
  \centering
  \includegraphics[width=0.9\columnwidth]{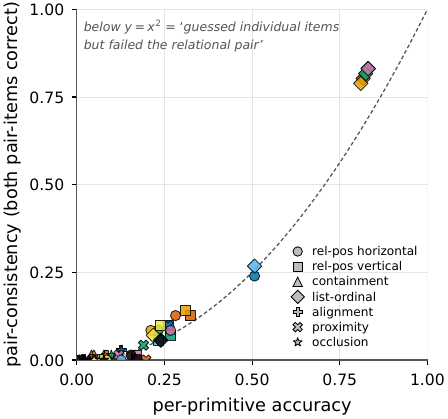}
  \caption{Pair-consistency vs.\ raw accuracy, per model per primitive. Independent guessing would place each point near $y = x^{2}$ (dashed). Almost all points lie \emph{below} this curve, the \textsc{What's-Up} signature of relational failure.}
  \label{fig:pair}
\end{figure}

\subsection{Pair-consistency: a relational failure signature}
\label{subsec:pair-consistency}
Independent draws would put pair-consistency at $y = x^{2}$ (accuracy squared). Figure~\ref{fig:pair} shows it consistently \emph{below} that curve, the relational-failure signature of \citet{kamath2023whatsup}. Within a model, the two pair members are negatively correlated: being right on the ``left of'' twin makes the model more likely than chance to be wrong on the ``right of'' twin. The relation word, not the picture, is what confuses the model (Appendix~\ref{app:pair-consistency}). Eight representative minimal-pair failures are analyzed in Appendix~\ref{app:qualitative}.


\begin{figure}[ht]
  \centering
  \includegraphics[width=0.9\columnwidth]{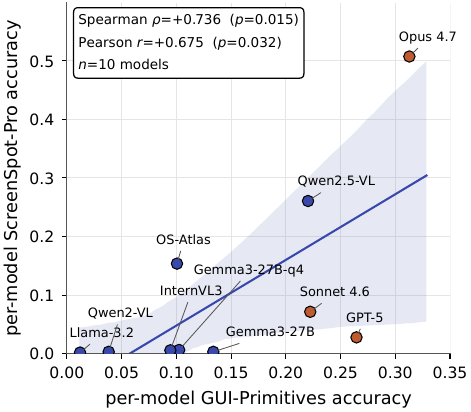}
  \caption{Per-model \bench{} accuracy ($n_{\text{models}} = 10$)
    against ScreenSpot-Pro grounding accuracy. Spearman
    $\rho = +0.736$ ($p = 0.015$).}
  \label{fig:scatter}
\end{figure}

\section{Primitive Competence Predicts Grounding}
\label{sec:regression}
A diagnostic is only useful if it predicts a downstream property of interest. We test this on ScreenSpot-Pro \citep{li2025screenspotpro}, the standard high-resolution GUI-grounding benchmark, which we ran on the ten models for which we have both \bench{} and SS-Pro predictions. Figure~\ref{fig:scatter} shows the per-model relationship. Spearman $\rho = +0.736$ ($p = 0.015$, $n = 10$); the correlation holds inside the real-screenshot subset ($\rho = +0.705$, $p = 0.023$) and inside the synthetic subset ($\rho = +0.760$, $p = 0.011$). A leave-one-model-out sensitivity analysis gives a Spearman in $[+0.62, +0.86]$; no single model drives the result, though dropping the strongest or weakest raises $p$ to $\approx 0.07$. An item-level logistic regression with model fixed effects (details in Appendix~\ref{app:regression}) and a log target-area control reaches pseudo-$R^{2} = 0.404$ on $n = 15{,}810$ pairs and corroborates the joint association; per-primitive coefficients are small ($|\text{coef}| \le 0.07$) and fragile, so we claim only the model-level effect (full forest plot in Appendix~\ref{app:forest}).

\section{Training-Free Interventions}
\label{sec:interventions}
If models lack the elementary spatial skills GUI grounding needs, can inference-time scaffolding put them back? We test three interventions. \textbf{SoM} \citep{yang2023setofmark} overlays numbered marks on candidate elements; the model names the mark, trading pixel regression for symbolic selection. \textbf{Primitive-aware CoT} prepends a short decomposition prompt that names the relation explicitly, probing whether the failure is a reasoning sequence rather than perception. \textbf{Activation steering} follows SteerVLM \citep{sivakumar2025steervlm}: we compute mean-difference vectors between correctly- and incorrectly-grounded items on a 200-item held-out contrast set, then add $\alpha \mathbf{v}_{\ell}$ to the residual stream at decoder layers $\ell \in \{12, \dots, 19\}$ of Qwen2.5-VL-7B at inference, with $\alpha = 4.0$ chosen as the largest value from $\{1, 2, 4, 6\}$ preserving fluency. As a positive control, the same pipeline with a verbosity-contrast vector shifts output length as expected, so the machinery is functional.

\begin{figure}[ht]
  \centering
  \includegraphics[width=0.9\columnwidth]{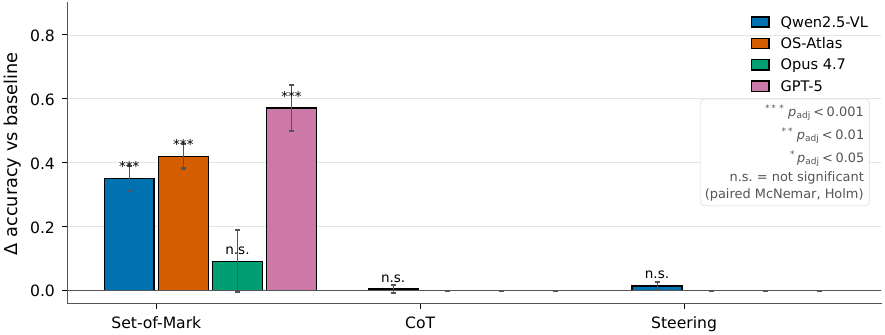}
  \caption{Intervention deltas vs.\ each model's baseline, with Holm-adjusted significance (paired McNemar across primitives). SoM recovers 35--57 points on three of four tested models; the only marginal case is Claude Opus 4.7, which already has the highest baseline. CoT and activation steering yield no significant change.}
  \label{fig:iv}
\end{figure}

\subsection{Oracle two-candidate marking}
\label{subsec:som}
Figure~\ref{fig:iv} reports intervention deltas with significance. Oracle two-candidate marking overlays marks on exactly the target and distractor, so the baseline and marked conditions are not identical-task comparisons: marking removes the candidate-search burden that dominates baseline error (\S\ref{subsec:perprim}). It is the only intervention producing large, statistically significant gains: $+57.1$ pt on \textbf{GPT-5} ($30 \to 87\%$, $p < 10^{-50}$ the highest absolute accuracy any model reaches under any condition), $+42.1$ pt on \textbf{OS-Atlas-Base-7B} ($10 \to 52\%$, $p < 10^{-50}$), $+35.1$ pt on \textbf{Qwen2.5-VL-7B} ($22 \to 57\%$, $p < 10^{-50}$), and a marginal $+9.2$ pt on \textbf{Claude Opus 4.7} ($31 \to 40\%$, Holm-adjusted $p = 0.08$, not significant). Across the four evaluated models, the observed gain is inversely ordered with baseline accuracy; given four points, we treat this as descriptive rather than a general relationship.

\begin{figure}[ht]
  \centering
  \includegraphics[width=0.9\columnwidth]{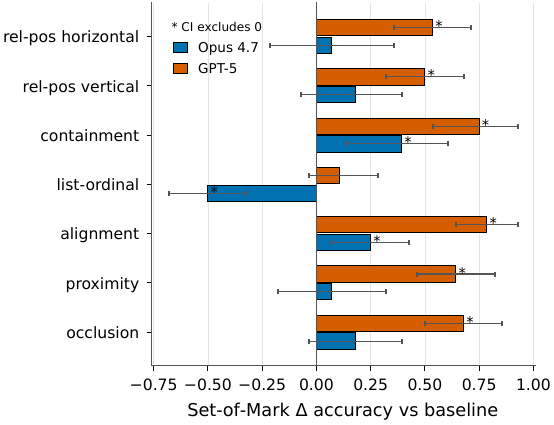}
  \caption{Per-primitive SoM delta on Claude Opus 4.7 and GPT-5. Stars mark intervals excluding zero. On GPT-5, six of seven primitives gain at least 40 points.}
  \label{fig:som-prim}
\end{figure}

\paragraph{Per-primitive pattern.}
Figure~\ref{fig:som-prim} breaks the gain down by primitive on the two closed APIs we ran with SoM. Six of seven primitives gain at least 40 points on GPT-5; the only primitive that \emph{regresses} (by $-32.4$ points) is \texttt{list-ordinal} on Qwen2.5 VL, which the model had already saturated at 81\% baseline. Converting an already-solved task into mark selection introduces an indirect cost that can reduce accuracy on primitives the model has already saturated.

\subsection{CoT and activation steering: negative results}
\label{subsec:negatives}
We report two negative results with the same effort as the positive one (Appendix~\ref{app:negatives}). Primitive-aware CoT on Qwen2.5-VL-7B produced $\Delta = +0.5$~pt ($p = 0.51$, n.s.); the model already names the relation correctly in its intermediate reasoning, then still emits the wrong coordinate. Activation steering produced $\Delta = +1.5$~pt ($p = 0.066$, n.s.), with no configuration in the grid crossing the Holm threshold. The one intervention that works bypasses the spatial skill it should test; the two that probe internal representation directly do not move the metric. We read this as evidence that the failure lies closer to perception (visual encoder, cross-modal binding) than to language-side reasoning or residual-stream representation.

\section{Discussion}
\label{sec:discussion}
Accuracy does not increase monotonically with nominal model scale in the evaluated set. The strongest proprietary models achieve only 24--31\% accuracy. Spatial grounding in GUI screenshots remains a persistent weakness of current vision-language systems. Oracle two-candidate marking provides the largest training-free improvement we observe, but it supplies the candidate set rather than repairing the skill gap, and is therefore a diagnostic upper bound rather than a solution. By marking candidate elements, SoM shifts part of the relational grounding problem into symbolic mark selection. Its effectiveness also depends on the quality of the upstream detector, whose failure modes are discussed in Appendix~\ref{app:som-failures}. Candidate-level analysis (Appendix~\ref{app:candidate}) indicates that below-reference accuracy is primarily attributable to predictions outside both candidate regions (60--92\% of predictions) rather than to systematic selection of the contrastive element; within-region selection follows the relation word on all primitives except containment and occlusion, where it is statistically indistinguishable from $0.50$. The associated layout-frequency prior is directional: predictions fall within a candidate region 2.7$\times$ more often when the target is the left element, and within-region accuracy is 0.972 under ``left of'' versus 0.717 under ``right of'' (Appendix~\ref{app:sensitivity}). We report this prior as a hypothesis for the source of the directional asymmetry rather than as a causal claim. The same decomposition is consistent with the magnitude of the SoM gain: SoM supplies candidate regions directly, addressing the candidate-localization component that accounts for most baseline errors. Appendix~\ref{app:errors} presents six recurring error types with representative examples. Finally, the failure pattern mirrors relational errors observed in \textsc{What's-Up} for natural images~\citep{kamath2023whatsup}. This suggests that controlled minimal-pair diagnostics are essential for evaluating spatial competence across visual modalities, including GUI grounding.

\section{Limitations}
\label{sec:limits}
This work has several limitations (Appendix~\ref{app:limits}). \textbf{(i)} Containment and occlusion appear only in the synthetic arm of \bench{} (list-ordinal includes a 24-item real slice), since real GUI datasets lack scalable parent--child or overlay--target metadata for minimal-pair mining; findings on these primitives therefore have limited external validity, and results are reported separately by source arm. This scoping bounds the candidate-level finding of Appendix~\ref{app:candidate}: the absence of a measurable relation-word signal on containment and occlusion is established on synthetic screenshots only. \textbf{(ii)} After model fixed effects absorb between-model variance, per-primitive coefficients in our item-level regression are small ($|\text{coef}| \le 0.07$); we claim only the joint model-level association and report the forest plot (Appendix~\ref{app:forest}) for transparency. \textbf{(iii)} We stop at ScreenSpot-Pro; end-to-end OSWorld task success \citep{xie2024osworld, osworldverified2025} remains future work. \textbf{(iv)} \bench{} v1 covers static pre-action grounding for English click targets only; planned extensions include drag-target prediction, bidirectional and CJK layouts, an OSWorld pre-action slice, and frozen snapshots of shifting closed APIs (Appendix~\ref{app:models}). \textbf{(v)} SoM acts as a scaffold by partially converting relational grounding into symbolic mark selection; our setup overlays at most two candidate marks, so gains may not generalize to realistic multi-candidate scenes, and deployment performance will inherit upstream detector failures. \textbf{(vi)} Point-in-box accuracy is binary and coarse: our loose metric (within $2\times$ target diagonal) partly separates sub-pixel imprecision from element misidentification, but neither metric captures structured spatial errors (e.g., anchor-collapse or direction inversion), which we analyze qualitatively in Appendix~\ref{app:errors}. Llama-3.2-11B-Vision emits a default coordinate on $\approx 42\%$ of items (1.2\% accuracy), consistent with ScreenSpot-Pro reports of $<\!2\%$ for generalist VLMs.

\section*{Ethical Considerations}
\label{sec:ethics}

Annotators were paid above the local hourly minimum and provided informed consent (Appendix~\ref{app:human-annotation}). UI-Vision screenshots remain under their original license; we redistribute only derived annotations, not the screenshots themselves. \bench{} is intended to support safer GUI-agent development by exposing failures in elementary spatial grounding before deployment.

\bibliography{main}

\appendix

\section{Full Per-Model Results}
\label{app:full-results}
\begin{figure*}[ht]
  \centering
  \includegraphics[width=0.92\textwidth]{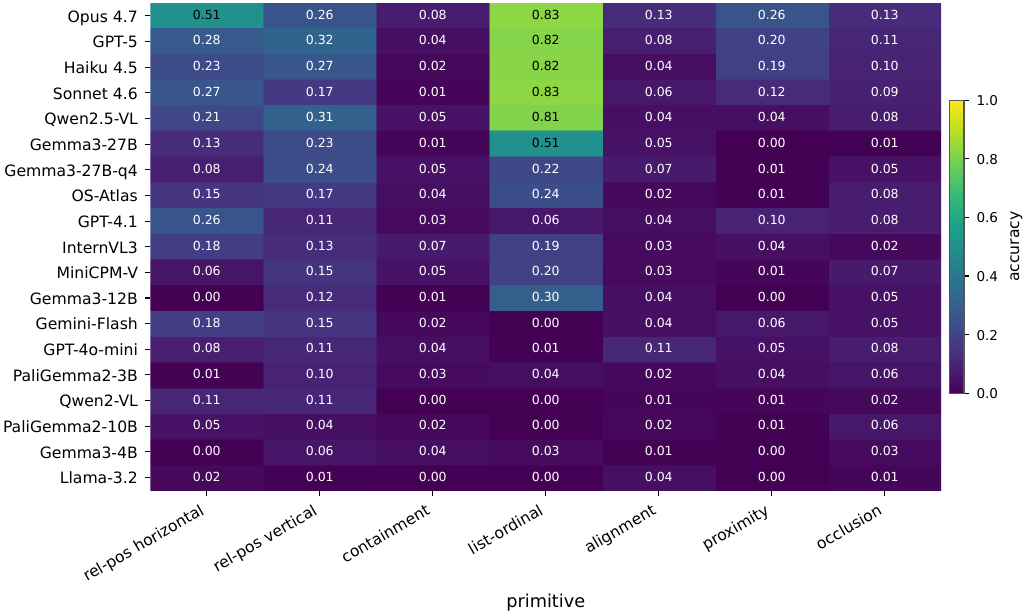}
  \caption{Per-model per-primitive accuracy heatmap, all 19 models. The horizontal band at $\sim$0.80 on \texttt{list-ordinal} marks the only primitive any model solves reliably.}
  \label{fig:heatmap}
\end{figure*}

Table~\ref{tab:full19} reports overall accuracy for all 19 models on the full benchmark, the 196-item human-verified core, and the 185-item human-clean core. Per-primitive accuracies with bootstrap 95\% CIs for every cell are released alongside the artifact in \texttt{runs/diagnostic/analysis.json}. A model-by-primitive accuracy heatmap is shown in Figure~\ref{fig:heatmap}.

\begin{table}[ht]
\small
\centering
\caption{Full benchmark accuracy for all 19 evaluated models on three splits: the full $n{=}994$ benchmark (\textbf{Full}), the human-verified $n{=}196$ subset (\textbf{Core}), and the cleaned $n{=}185$ subset after removing items judged invalid by a majority of the five annotators (\textbf{Clean}). Closed models were not run on the full split because of API cost; the Core column reports every model that was. Per-primitive accuracies with bootstrap intervals are released in the artifact.}
\setlength{\tabcolsep}{4pt}
\begin{tabular}{@{}lccc@{}}
\toprule
\textbf{Model} & \textbf{Full} & \textbf{Core} & \textbf{Clean} \\
\midrule
Claude Opus 4.7              & 0.313 & 0.311 & 0.324 \\
GPT-5 (reasoning)            & 0.265 & 0.291 & 0.297 \\
Claude Haiku 4.5             & 0.239 & 0.250 & 0.249 \\
Claude Sonnet 4.6            & 0.222 & 0.260 & 0.270 \\
Qwen2.5-VL-7B-Instruct       & 0.220 & 0.245 & 0.249 \\
Gemma 3-27B-it (HF)          & 0.134 & ---   & 0.157 \\
gemma3:27b (ollama, Q4)      & 0.103 & ---   & 0.125 \\
OS-Atlas-Base-7B             & 0.101 & 0.107 & 0.108 \\
GPT-4.1                      & 0.096 & ---   & 0.108 \\
InternVL3-8B                 & 0.095 & ---   & 0.097 \\
MiniCPM-V (ollama)           & 0.081 & ---   & 0.092 \\
Gemma 3-12B-it (HF)          & 0.074 & ---   & 0.087 \\
Gemini 3.1 Flash Lite        & 0.073 & 0.077 & 0.076 \\
GPT-4o-mini                  & 0.068 & 0.077 & 0.076 \\
PaliGemma 2-3B-mix-448       & 0.041 & ---   & 0.054 \\
Qwen2-VL-7B-Instruct         & 0.038 & 0.046 & 0.049 \\
PaliGemma 2-10B-mix-448      & 0.029 & ---   & 0.038 \\
Gemma 3-4B-it (HF)           & 0.022 & ---   & 0.027 \\
Llama-3.2-11B-Vision         & 0.012 & 0.005 & 0.005 \\
\bottomrule
\end{tabular}
\label{tab:full19}
\end{table}

\section{Models, Versions, and Endpoints}
\label{app:models}
Table~\ref{tab:models} lists the exact model identifiers used in this paper, together with the endpoint each was queried through. Closed models were accessed via their official APIs in May 2026; open models were run from Hugging Face checkpoints. We pin identifiers here so later runs can be compared against the snapshot reported in this
paper.

\begin{table*}[ht]
\small
\centering
\caption{Model identifiers used in this paper. Closed models were queried via official APIs in May 2026; open models were run from Hugging Face checkpoints with greedy decoding ($T = 0$) and a fixed seed.}
\label{tab:models}
\setlength{\tabcolsep}{6pt}
\begin{tabular}{@{}lll@{}}
\toprule
\textbf{Model} & \textbf{Identifier} & \textbf{Family / Endpoint} \\
\midrule
Claude Opus 4.7          & \texttt{claude-opus-4-7}                          & Anthropic API \\
Claude Sonnet 4.6        & \texttt{claude-sonnet-4-6}                        & Anthropic API \\
Claude Haiku 4.5         & \texttt{claude-haiku-4-5}                         & Anthropic API \\
GPT-5 (reasoning)        & \texttt{gpt-5}                                    & OpenAI API \\
GPT-4.1                  & \texttt{gpt-4.1}                                  & OpenAI API \\
GPT-4o-mini              & \texttt{gpt-4o-mini}                              & OpenAI API \\
Gemini 3.1 Flash Lite    & \texttt{gemini-3.1-flash-lite}                    & Google API \\
\midrule
Qwen2.5-VL-7B-Instruct   & \texttt{Qwen/Qwen2.5-VL-7B-Instruct}              & Hugging Face \\
Qwen2-VL-7B-Instruct     & \texttt{Qwen/Qwen2-VL-7B-Instruct}                & Hugging Face \\
OS-Atlas-Base-7B         & \texttt{OS-Copilot/OS-Atlas-Base-7B}              & Hugging Face \\
InternVL3-8B             & \texttt{OpenGVLab/InternVL3-8B}                   & Hugging Face \\
Llama-3.2-11B-Vision     & \texttt{meta-llama/Llama-3.2-11B-Vision-Instruct} & Hugging Face \\
Gemma 3-27B-it           & \texttt{google/gemma-3-27b-it}                    & Hugging Face \\
Gemma 3-12B-it           & \texttt{google/gemma-3-12b-it}                    & Hugging Face \\
Gemma 3-4B-it            & \texttt{google/gemma-3-4b-it}                     & Hugging Face \\
PaliGemma 2-10B-mix-448  & \texttt{google/paligemma2-10b-mix-448}            & Hugging Face \\
PaliGemma 2-3B-mix-448   & \texttt{google/paligemma2-3b-mix-448}             & Hugging Face \\
MiniCPM-V (ollama)       & \texttt{minicpm-v}                                & Ollama (Q4) \\
gemma3:27b (ollama)      & \texttt{gemma3:27b}                               & Ollama (Q4) \\
\bottomrule
\end{tabular}
\end{table*}

\section{Real vs.\ Synthetic: Strict and Loose Accuracy}
\label{app:realsyn-table}

Table~\ref{tab:realsyn} reports point-in-box (strict) and within-$2{\times}$-diagonal
(loose) accuracy split by source for seven representative models. The zero strict accuracy on UI-Vision real screenshots is not a target-size artifact: loose accuracy stays low ($6$--$23\%$), so the model picks the wrong element rather than missing the right one by a few pixels.

\begin{table}[ht]
\small
\centering
\caption{Strict (point-in-box) vs.\ loose (within $2\times$ target-diagonal) accuracy, split by image source. In real desktop screenshots, loose accuracy stays low even when the prediction is allowed to miss by a generous margin: failure is element misidentification, not sub-pixel imprecision.}
\label{tab:realsyn}
\setlength{\tabcolsep}{3.6pt}
\begin{tabular}{@{}lcccc@{}}
\toprule
& \multicolumn{2}{c}{\textbf{Real (UI-Vision)}} &
  \multicolumn{2}{c}{\textbf{Synthetic}} \\
\cmidrule(lr){2-3}\cmidrule(lr){4-5}
\textbf{Model} & strict & loose & strict & loose \\
\midrule
Claude Haiku 4.5         & 0.000 & 0.230 & 0.363 & 0.763 \\
Qwen2.5-VL-7B            & 0.028 & 0.234 & 0.300 & 0.714 \\
OS-Atlas-Base-7B         & 0.000 & 0.162 & 0.142 & 0.724 \\
Gemini 3.1 Flash Lite    & 0.000 & 0.115 & 0.111 & 0.741 \\
GPT-4o-mini              & 0.000 & 0.033 & 0.111 & 0.733 \\
Qwen2-VL-7B              & 0.000 & 0.066 & 0.054 & 0.693 \\
Llama-3.2-11B-Vision     & 0.000 & 0.062 & 0.017 & 0.616 \\
\bottomrule
\end{tabular}
\end{table}

\section{Item-Level Regression Forest Plot}
\label{app:forest}
\begin{figure}[ht]
  \centering
  \includegraphics[width=0.95\columnwidth]{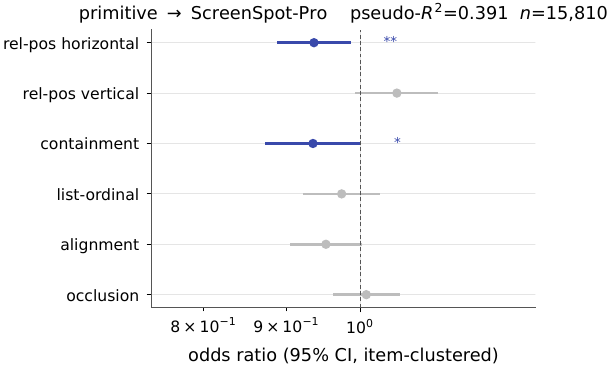}
  \caption{Item-level regression forest plot
    ($n = 15{,}810$, pseudo-$R^{2} = 0.404$).}
  \label{fig:forest}
\end{figure}

Figure~\ref{fig:forest} shows the item-level logistic regression of ScreenSpot-Pro success on per-primitive \bench{} competence ($n = 15{,}810$, pseudo-$R^{2} = 0.404$), with model fixed effects, a log target-area control, and item-clustered standard errors. After model effects absorb between-model variance, individual primitive coefficients are small ($|\text{coef}| \le 0.07$) and only
two cross the Holm-corrected threshold, so we report this plot for transparency only. \textbf{The headline number is the model-level Spearman in Fig.~\ref{fig:scatter} of the main paper, not the
per-primitive coefficients.}

\section{Closed-API Coordinate Rescaling}
\label{app:rescale}
All three closed-vision APIs we tested apply server-side image preprocessing whose output is not visible to the caller. The predicted click coordinates are returned in the \emph{downsampled} frame rather than the original pixel space, silently misaligning the prediction with the ground-truth bounding box.

\paragraph{Detection.}
On the first GPT-5 ScreenSpot-Pro run, we observed predictions like $(662, 541)$ on items whose ground-truth boxes were at $(1774, 1586, 2113, 1618)$ in a $3840 \times 2160$ image. The prediction was internally consistent with a $1024$-side downsample; the ground truth was in the original frame.

\paragraph{Mitigation.}
Our wrapper now resizes every image to a known maximum long side ($\le 1500$ px) before submission, then rescales the predicted coordinate back to the original space using the recorded resize factor. Where the provider does further internal tiling (notably OpenAI), we additionally validate by re-querying on the already-downsized image and confirming the rescale factor is $\approx 1.0$. Records that required rescaling: 42/196 Claude Haiku
diagnostic, 43/196 Claude Opus diagnostic, 219/994 Claude Sonnet diagnostic, 1{,}297/1{,}581 Claude Sonnet ScreenSpot-Pro, and $\approx 1500/1581$ GPT-5 ScreenSpot-Pro. The 284 Claude Sonnet ScreenSpot-Pro items that originally errored with HTTP~400 (``image $> 5$~MB'') were retried under the new wrapper, and the final run is clean.

\paragraph{Recommendation.}
We recommend that future closed-VLM evaluations report whether their wrapper handles provider-side downsampling. Otherwise, ScreenSpot-Pro's strict accuracy on high-resolution professional applications is silently and severely under-reported.

\section{Human Annotation Protocol}
\label{app:human-annotation}
Five annotators independently labeled the 196-item stratified subset. Each annotator answered two questions per item.

\paragraph{Q1: well-formedness.}
\textit{Is the instruction unambiguous on this screenshot?} The annotator answered \texttt{good} or \texttt{bad}. An item was kept only if at least 3 of 5 annotators marked it as well-formed. This dropped 11 of 196 items, leaving the 185-item \emph{human-clean} core.

\paragraph{Q2: correct target.}
\textit{Which of the two candidate boxes does the instruction identify?} The annotator answered \texttt{t} (the target box) or \texttt{d} (the distractor box). Human accuracy is the fraction of
items on which the majority answer was \texttt{t}, computed on the human-clean core only.

\paragraph{Agreement.}
We report Fleiss $\kappa$ for both questions on the full 196-item subset. Fleiss $\kappa_{Q1} = 0.942$ (well-formedness, ``almost perfect''); Fleiss $\kappa_{Q2} = 0.787$ (target, ``substantial''). The pairwise $\kappa$ matrix is released with the artifact.

\paragraph{Compensation and consent.}
Annotators were paid above the local hourly minimum, were told the purpose of the labeling (academic publication and public benchmark release), and could opt out at any point. No personally-identifying information was collected.

\section{Synthetic-Stimulus Generation}
\label{app:synthetic}
The synthetic arm of \bench{} renders GUI-like screenshots with a fixed layout language: a top toolbar of buttons, a left sidebar of list items, a central grid of tiles, and an optional dialog overlay. Element positions are sampled uniformly within constrained pixel ranges; the seed is fixed to ensure reproducibility of the corpus. For each primitive, a separate generator constructs minimal pairs by selecting an anchor and the two candidate elements that satisfy the contrastive relation. Quality gates in the released code verify that both members of every pair use the same screenshot, have non-empty disjoint targets, and pass per-primitive coverage checks before items enter the benchmark.

\section{Intervention Implementation Details}
\label{app:interventions}

\paragraph{SoM.}
We render mark IDs on the candidate boxes (two per item: target and distractor) with a deterministic ID shuffle keyed by the item ID so the correct answer is not always ``1''. The model is asked to respond with the mark number to click, and we map the response back to the box center for evaluation.

\paragraph{Primitive-aware CoT.}
We prepend a one-paragraph decomposition prompt that names the relation explicitly and asks for a chain of reasoning before the click. Per-primitive scaffolds are released in the artifact.

\paragraph{Activation steering.}
We hook the decoder layers of Qwen2.5-VL-7B at layers
$\{12, \dots, 19\}$ and add $\alpha \cdot \mathbf{v}_{\ell}$ to the residual stream at each layer, where $\mathbf{v}_{\ell}$ is the mean-difference vector between hidden states on correctly- and incorrectly-grounded baseline items at that layer. We searched $\alpha \in \{1, 2, 4, 6\}$; $\alpha = 4$ was the maximum that left sentence-level fluency intact.

\section{SoM Failure Modes}
\label{app:som-failures}
SoM replaces the original grounding task with a mark-selection task. The substitution is not perfect, and the failure modes are worth listing for downstream users.

\textbf{Saturated primitive regression.} On Qwen2.5-VL-7B, \texttt{list-ordinal} accuracy drops by 32.4 points under SoM. The model had already learned to count list items by their on-screen order; converting the same task to ``select mark X'' adds an indirection that the model handles less well than the original.

\textbf{Mark placement collisions.} On highly crowded screenshots, two marks placed at nearby element centers can occlude each other, or the model can confuse them. We did not observe this on the small candidate sets used in our benchmark (typically 2 marks per item), but it is a known issue at scale.

\textbf{Closed-API consistency.} On the human-verified core, Claude Opus 4.7 gained only $+9.2$ points under SoM, in contrast to GPT-5's $+57.1$ points. This 47-point gap on the same items under the same intervention suggests that Opus already grounds the unmarked image with the same logic SoM provides, while GPT-5 was genuinely held back by an aspect of the task that mark selection removes.

\section{Details of the GUI-Primitives Benchmark}

\subsection{Details of minimal-pair construction}
\label{app:minpair}
Every item in \bench{} appears as one half of a contrastive pair. A pair shares one screenshot and one anchor element; the relation expression changes between the two members so that the correct target moves to the other designated candidate. In 253 of 497 pairs, the surrounding instruction template also varies lexically in verb, head noun, and anchor quoting; this variation is not confounded with the relation term (Appendix~\ref{app:sources}). Figure~\ref{fig:teaser} shows a real example: in (A) the click should land to the right of the \texttt{break debugger} icon; in (B) the same screenshot is paired with \texttt{left of}, and the correct target is the icon on the other side. The construction controls for screenshot-specific salience and fixed answer preferences: a model that always selects the most central, salient, or frequently labeled element scores correctly on one twin and incorrectly on the other, so its pair-level score is zero. We measure this directly with \emph{pair-consistency}: the fraction of pairs on which both members are correct (\S\ref{subsec:pair-consistency}).

\subsection{Details of sources: real and synthetic screenshots}
\label{app:sources}

\bench{} draws screenshots from two sources, both reported separately.

\textbf{Real desktop screenshots (n = 290 items).} We use the element-grounding annotations from UI-Vision \citep{uivision} as the source corpus for real GUIs, covering Web, MacOS, VS Code, Office,
and various professional applications. Items synthesized from this source cover the four primitives whose annotations are unambiguously
recoverable from element labels: horizontal relative position, vertical relative position, alignment, and proximity.

\textbf{Controlled synthetic stimuli (n = 704 items).} The remaining items are rendered from a procedural generator that places elements
in known positions, with controlled distractors and a fixed layout language. This is the only way to obtain \emph{containment}, \emph{list ordinal}, and \emph{occlusion} items in clean form, because those three primitives require a structured parent--child or overlay--target relationships that real screenshot annotations do not ship at scale.

We disclose the split openly. The synthetic arm is a controlled-stimulus arm in the \textsc{What's-Up} tradition, where the hardest relations are likewise generated under controlled rendering. All real vs. synthetic numbers are reported separately throughout the paper.

\subsection{Details of human verification}
\label{app:verification}
Five annotators independently judged a 196-item stratified subset (\textit{human-verified core}). For each item, the annotator answered two questions: is the instruction well-formed and unambiguous on the given screenshot, and which of the two candidate boxes is the correct target. Inter-annotator agreement is high. On Q1 (well-formedness), Fleiss $\kappa = 0.942$ (``almost perfect'' on the Landis--Koch scale, \citealp{landis1977measurement}). On Q2 (correct target), Fleiss $\kappa = 0.787$ (``substantial''). Eleven items were judged invalid by the majority and were dropped, leaving 185 items in the \textit{human-clean core}.

\begin{figure*}[h]
  \centering
  \includegraphics[width=0.7\textwidth]{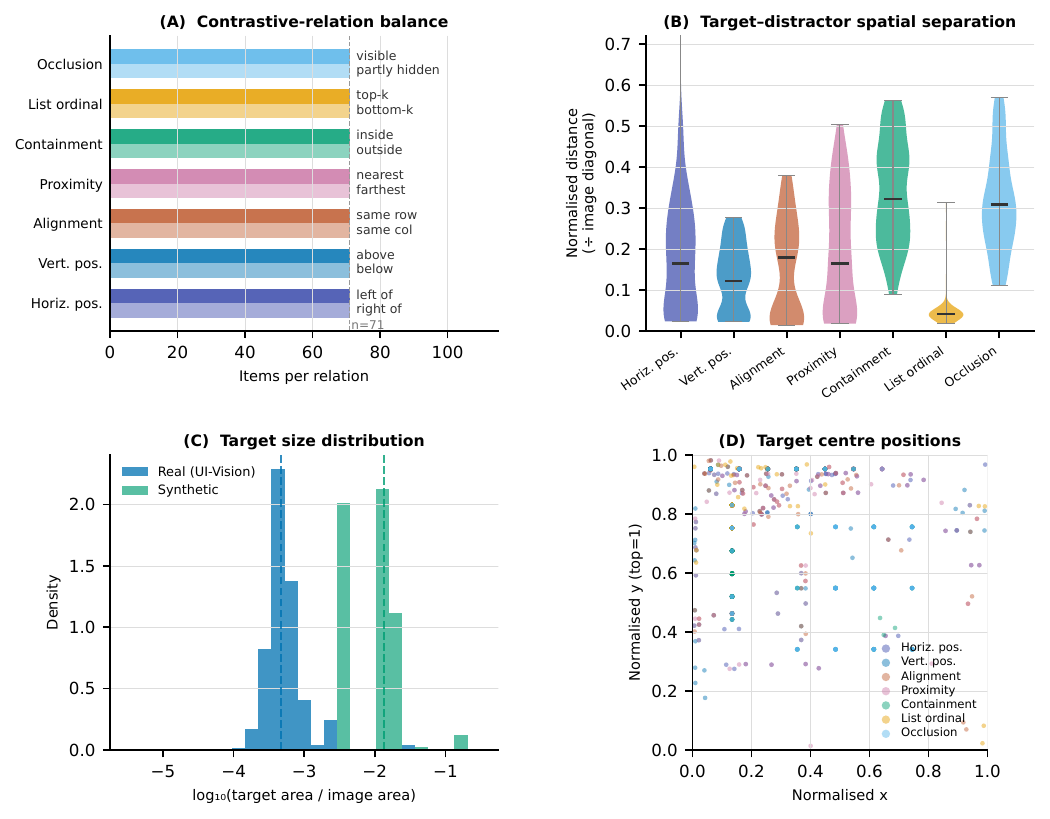}
  \caption{\textbf{Benchmark distribution analysis.}
    \textbf{(A)} Contrastive-relation balance: every relation is represented by exactly 71 items per side, ruling out relation-prior exploitation.
    \textbf{(B)} Target--distractor spatial separation (normalized by image diagonal) per primitive; violin width encodes density.
    \textbf{(C)} Target size distribution (log-scale) by source arm; dashed lines are per-source medians.
    \textbf{(D)} Target centre positions (normalized to $[0,1]^{2}$, $y{=}1$ at top), coloured by primitive; targets span the full screen area.}
  \label{fig:distribution}
\end{figure*}

\section{Benchmark Distribution and Scenario Diversity}
\label{app:distribution}

Figure~\ref{fig:distribution} summarises the distributional properties of \bench{} across four dimensions.

\paragraph{Relation balance (Panel A).}
Every contrastive relation is represented by exactly 71 items per side per primitive, a strict 50/50 split enforced by the minimal-pair miner.
This guarantees that a model exploiting a relation-frequency prior (e.g., always predicting \texttt{left}) cannot achieve better-than-reference-level accuracy at the primitive level, and it rules out skewed-prior explanations for any observed performance gap.
Table~\ref{tab:composition} reports the per-primitive composition by source arm: 142 items per primitive; containment and occlusion are synthetic-only, and list-ordinal includes a 24-item real slice.

\begin{table}[ht]
\small
\centering
\caption{Per-primitive composition of \bench{} by source arm. Every primitive has 142 items, exactly 71 per relation word.}
\label{tab:composition}
\setlength{\tabcolsep}{4pt}
\begin{tabular}{@{}lrrr@{}}
\toprule
\textbf{Primitive} & \textbf{Real} & \textbf{Synthetic} & \textbf{Total} \\
\midrule
rel-pos-horizontal & 68 & 74  & 142 \\
rel-pos-vertical   & 48 & 94  & 142 \\
alignment          & 66 & 76  & 142 \\
proximity          & 84 & 58  & 142 \\
list-ordinal       & 24 & 118 & 142 \\
containment        & 0  & 142 & 142 \\
occlusion          & 0  & 142 & 142 \\
\midrule
Total              & 290 & 704 & 994 \\
\bottomrule
\end{tabular}
\end{table}

\paragraph{Target--distractor spatial separation (Panel B).}
Target--distractor distances (normalized by the image diagonal) span a wide range: median $0.18$, IQR $[0.04, 0.31]$.
\texttt{list-ordinal} items are tightly spaced (adjacent list rows, median $0.04$), while \texttt{containment} and \texttt{occlusion} items have the largest separations (median $0.32$--$0.31$).
This variation ensures that the benchmark exercises a range of spatial configurations rather than collapsing onto a single difficulty level.

\paragraph{Target size (Panel C).}
Real UI-Vision targets are considerably smaller than synthetic ones (median $0.05\%$ vs.\ $1.3\%$ of image area), consistent with the fine-grained element boxes in professional desktop applications. Both distributions span multiple orders of magnitude (log$_{10}$ range $\approx$3.5), providing diversity in target granularity within each arm.

\paragraph{Spatial coverage (Panel D).}
Target centers, normalized to the unit square, span the full screen area with no strong clustering, reflecting a mix of toolbar, sidebar, central panel, and dialog overlay targets across both sources. The slight top-left bias ($\bar{x}=0.29$, $\bar{y}=0.24$) reflects the prevalence of toolbar and sidebar elements in real GUI layouts rather than a construction artifact.

\section{Details of Models and Protocol}
\label{app:protocol}

\paragraph{Details of models.} We evaluate 19 vision-language models: seven proprietary models (Claude Opus 4.7, Claude Sonnet 4.6, Claude Haiku 4.5, GPT-5, GPT-4.1, GPT-4o-mini, Gemini 3.1 Flash Lite); three open Qwen-family models (Qwen2.5-VL-7B, Qwen2-VL-7B, OS-Atlas-Base-7B, which is GUI-tuned Qwen2-VL); InternVL3-8B; Meta Llama-3.2-11B-Vision; three sizes of Google Gemma 3 (4B, 12B, 27B, including an Ollama-Q4 copy); two PaliGemma~2 variants (3B, 10B); and MiniCPM-V. The full list and provenance are in Appendix~\ref{app:models}. We could not include Pixtral-12B and Idefics3-8B because their chat-template formats are incompatible with our generic wrapper; this is disclosed as a known scope limitation rather than silently omitted.

\paragraph{Details of prompting and decoding.}
All models use greedy decoding ($T = 0$) and a fixed seed. The grounding prompt is a single short instruction (``respond with only the pixel coordinate to click''). Per-family coordinate conventions (pixel, normalized $[0,1]$, normalized $[0,1000]$) are handled by a robust parser; the parser is in the released code, and its unit tests are bundled with the artifact.

\paragraph{Details of closed-API coordinate frames.}
A practical wrinkle complicated the closed-model runs. The three closed providers each downsample large screenshots server-side (Anthropic to $\le 1568$~px, OpenAI to roughly $1024$ on the long side, Gemini variable) and return coordinates in the downsampled frame. We pre-resize every image to a known long side and rescale the predicted coordinate back to the original pixel space. The rescale was needed on a large fraction of items (e.g., 1{,}297 of 1{,}581 Sonnet ScreenSpot-Pro records). Full details in Appendix~\ref{app:rescale}; we believe this is a methodological note worth sharing with the community.

\paragraph{Details of metrics.}
The primary metric, following ScreenSpot-Pro, is \emph{point-in-box} accuracy: a prediction $\hat{p} = (\hat{x}, \hat{y})$ is correct iff $\hat{x} \in [x_{1}, x_{2}]$ and $\hat{y} \in [y_{1}, y_{2}]$ for ground-truth box $B = [x_{1}, y_{1}, x_{2}, y_{2}]$ in the original pixel frame. \emph{Loose} accuracy (prediction within $2\times$ target diagonal of the box center) separates sub-pixel imprecision from element mis-identification on small icons. All accuracies carry bootstrap 95\% CIs (2{,}000 resamples). Paired comparisons use McNemar \citep{mcnemar1947note} tests with Holm--Bonferroni \citep{holm1979simple} correction across the seven primitives. All primitives are binary contrasts (reference level $0.50$; Table~\ref{tab:primitives}). We flag a cell as below the reference level only when the upper bootstrap bound falls below $0.50$; thus, a point estimate dipping below it due to sampling noise does not count. Because models emit unconstrained coordinates, $0.50$ is a forced-choice reference rather than the chance level of the prediction task (Appendix~\ref{app:candidate}). The below-reference flags persist under anchor-level and screenshot-level cluster resampling; the synthetic generator reuses anchor slots (containment: 4 distinct anchors; occlusion: 13; alignment: 34; proximity: 43, each over 142 items and 71 screenshots), and no clustered upper bound reaches $0.50$ for any of the 19 models on the four sub-reference primitives.

\section{Details of CoT and Activation Steering: Negative Results}
\label{app:negatives}
We report two negative results, with the same effort as the positive one.

Primitive-aware CoT on Qwen2.5-VL-7B produced $\Delta = +0.5$ points (Holm-adjusted $p = 0.51$, not significant). The decomposition prompt does not seem to address the limiting factor: the model correctly names the relation in its intermediate reasoning, yet still emits the wrong coordinate.

Activation steering with SteerVLM-style vectors on Qwen2.5-VL-7B produced $\Delta = +1.5$ points (Holm-adjusted $p = 0.066$, not significant). The grid search over layers and $\alpha$ did not yield a
configuration that crossed the Holm threshold. Either GUI grounding requires a different intervention point than the layers where SteerVLM has helped on linguistic tasks, or the residual-stream signal is genuinely weak, and a single contrastive vector is not enough to push the model across the decision boundary on minimal pairs.

\paragraph{Reading the intervention table.}
The picture is uneven on purpose. The one intervention that works, SoM, works by \emph{bypassing} the spatial skill it should
test, replacing it with a symbolic naming task. The two interventions that probe the model's internal representation directly do not move the metric. We read this as evidence that the failure is closer to perception (visual encoder, cross-modal binding) than to language-side reasoning or residual-stream representation.

\section{Details of Diagnostic Results}
\label{app:results}

\subsection{Details of shortcut controls validate the benchmark}
\label{app:controls}
A diagnostic is only as good as the shortcuts it forbids. Figure~\ref{fig:controls} reports three controls on Qwen2.5-VL-7B over the full 994-item benchmark.

In the \emph{text-only} condition, the instruction is paired with a blank canvas of the same size: accuracy falls from 22.0\% to 6.5\% ($\Delta = -15.5$~pt, McNemar $p = 2.8 \times 10^{-27}$). In the \emph{shuffled} condition, the screenshot is replaced with one from a different item of the same primitive: accuracy drops to 14.8\% ($\Delta = -7.2$~pt, $p = 1.3 \times 10^{-11}$). Both controls behave as a truly grounded model should: when the visual or the matching instruction is removed, the model can no longer solve the task.

The third control, \emph{heavy Gaussian blur} ($\sigma = 12$~px), drops accuracy to 16.7\% ($\Delta = -5.3$~pt, $p = 8.5\times 10^{-5}$). The drop is statistically significant but below our pre-registered ten-point threshold, so under our pre-registered criterion the blur control does not reach a strict pass. The result indicates that VLM grounding relies more on global layout than on fine-grained OCR of text labels, since removing high-frequency detail removes only a modest fraction of the signal.

\subsection{Details of overall accuracy and the human gap}
\label{app:overall}
Table~\ref{tab:overall} reports the headline numbers.
Figure~\ref{fig:gap} visualizes the gap from human performance. Three observations follow.

First, all models perform well below the annotator target-selection accuracy of 96.9\% on a two-candidate task, which is not matched to model point-in-box accuracy (\S\ref{sec:humanverif}). The strongest system, Claude Opus 4.7, reaches 31--32\%. The 65-point gap holds on the full benchmark and on the cleaned 185-item core alike.

Second, accuracy does not increase monotonically with nominal scale. The four strongest proprietary models (Claude Opus 4.7, GPT-5, Claude Sonnet 4.6, Claude Haiku 4.5) occupy the top of the table but cluster within a narrow 24--31\% range. Meanwhile, the 7B open Qwen2.5-VL-Instruct is statistically tied with the closed Claude Haiku 4.5 on the human-verified core (24.5\% vs.\ 25.0\%), a noteworthy parity finding given that Qwen2.5-VL is roughly two orders of magnitude smaller than the closed reference and runs on a single GPU.

Third, the rank order is stable. The same top group leads on the full 994-item benchmark, on the 196-item human-verified core, and on the LLM-judge-verified 100-item subset (Appendix~\ref{app:full-results}). This means the human gap is not an artifact of noisy items.

\subsection{Details of per-primitive structure of failure}
\label{app:perprim}
Figure~\ref{fig:perprim} unpacks the overall accuracy by primitive.
The picture is uneven and informative.

One primitive, \texttt{list-ordinal}, is well solved by the top five models (Claude family, GPT-5, and Qwen2.5-VL), all achieving $0.80$--$0.83$ against the $0.50$ two-candidate reference (Table~\ref{tab:primitives}). The shared task here is counting list positions, which the language model can do once it identifies the list.

The other six primitives are uniformly hard. On
\texttt{rel-pos-horizontal} Claude Opus 4.7 reaches $0.51$, at the reference level; every other model is below. On \texttt{rel-pos-vertical}, the best model is GPT-5 at $0.32$. On \texttt{containment}, \texttt{occlusion}, \texttt{alignment}, and \texttt{proximity}, \emph{the upper 95\% bootstrap bound for every model we test sits below the reference level} of $0.50$, including all four strongest proprietary models. Candidate-level analysis (Appendix~\ref{app:candidate}) indicates that this below-reference accuracy is primarily attributable to predictions outside both candidate regions (60--92\% of predictions) rather than to selection of the contrastive element: within-region selection favors the target at least as often as the distractor on every primitive, with containment and occlusion statistically indistinguishable from $0.50$ (0.548 / 0.554) and horizontal, vertical, and proximity at 0.89--0.90. The associated directional prior is discussed in \S\ref{sec:discussion}.

\subsection{Details of real versus synthetic}
\label{app:realsyn}
We separate two candidate explanations for the per-primitive results in \S\ref{subsec:perprim}: small real-GUI target size and the relative proportion of synthetic items. Table~\ref{tab:realsyn} reports strict and loose accuracy split by source.

On synthetic screenshots, models often predict near the correct element: loose accuracy is $0.60$--$0.76$ even when strict box-edge accuracy is substantially lower, indicating that most synthetic errors are sub-pixel imprecision rather than element misidentification. On UI-Vision real screenshots, strict accuracy is essentially zero for every model on every primitive, and \emph{loose} accuracy also stays low (6--23\%). Candidate-level analysis (Appendix~\ref{app:candidate}) shows that on the real arm, 96\% of predictions fall outside both candidate regions, while among predictions inside a candidate region, the target is selected on 0.75 pooled. Real-screenshot accuracy is therefore primarily attributable to candidate localization, with target size a contributing factor, rather than to inverted relational interpretation.

\subsection{Details of pair-consistency: a relational failure signature}
\label{app:pair-consistency} 
If a model produced independent draws on the two members of a pair, pair-consistency would equal $y = x^{2}$ (accuracy squared). Figure~\ref{fig:pair} shows that pair-consistency is consistently \emph{lower} than that $y = x^{2}$ curve. This pattern, called the relational-failure signature in \citet{kamath2023whatsup}, means that within a model, the two pair members are negatively correlated: when the model is right on the ``left of'' twin, it is more likely than chance to be wrong on the ``right of'' twin. The information that confuses the model is the relation word, not the picture.

\section{Details of Primitive Competence Predicts Grounding}
\label{app:regression}
A diagnostic is only useful if it predicts a downstream property of interest. We test this on ScreenSpot-Pro \citep{li2025screenspotpro}, the standard high-resolution GUI-grounding benchmark, and we run it on the ten models for which we have both \bench{} and SS-Pro predictions.

Figure~\ref{fig:scatter} shows the per-model relationship. Spearman $\rho = +0.736$ ($p = 0.015$, $n = 10$). The correlation holds within the real-screenshot subset ($\rho = +0.705$, $p = 0.023$) and within the synthetic subset ($\rho = +0.760$, $p = 0.011$), so it is not an artifact of any single item type. A leave-one-model-out sensitivity analysis gives a Spearman in $[+0.62, +0.86]$, so no single model drives the result; dropping the strongest or weakest does raise the $p$-value to $\approx 0.07$. We also report a wide non-parametric bootstrap CI ($[+0.11, +0.98]$, $n = 10$) for transparency rather than as a precision claim. This is the load-bearing result of this
section.

\paragraph{Item-level corroboration.}
We also fit an item-level logistic regression of ScreenSpot-Pro success on each item, using the model's per-primitive accuracy profile, model fixed effects, a log target-area control, and item-clustered standard errors. The model achieves pseudo-$R^{2} = 0.404$ on $n = 15{,}810$ model-item pairs. The forest plot of per-primitive coefficients is in Figure~\ref{fig:forest}.

We urge caution on the per-primitive coefficients. After the model fixed effects absorb between-model variance, the residual between-primitive variance is small (all $|\text{coef}| \le 0.07$), and only two coefficients are individually significant at the Holm-corrected level. With four distinct open models contributing to ScreenSpot-Pro data, the competence vectors are correlated across primitives, making per-primitive identification fragile. We therefore claim only the joint, model-level association, which is strong, and explicitly do not claim that competence at any particular primitive causes downstream success.

\begin{figure*}[h]
  \centering
  \includegraphics[width=0.7\textwidth]{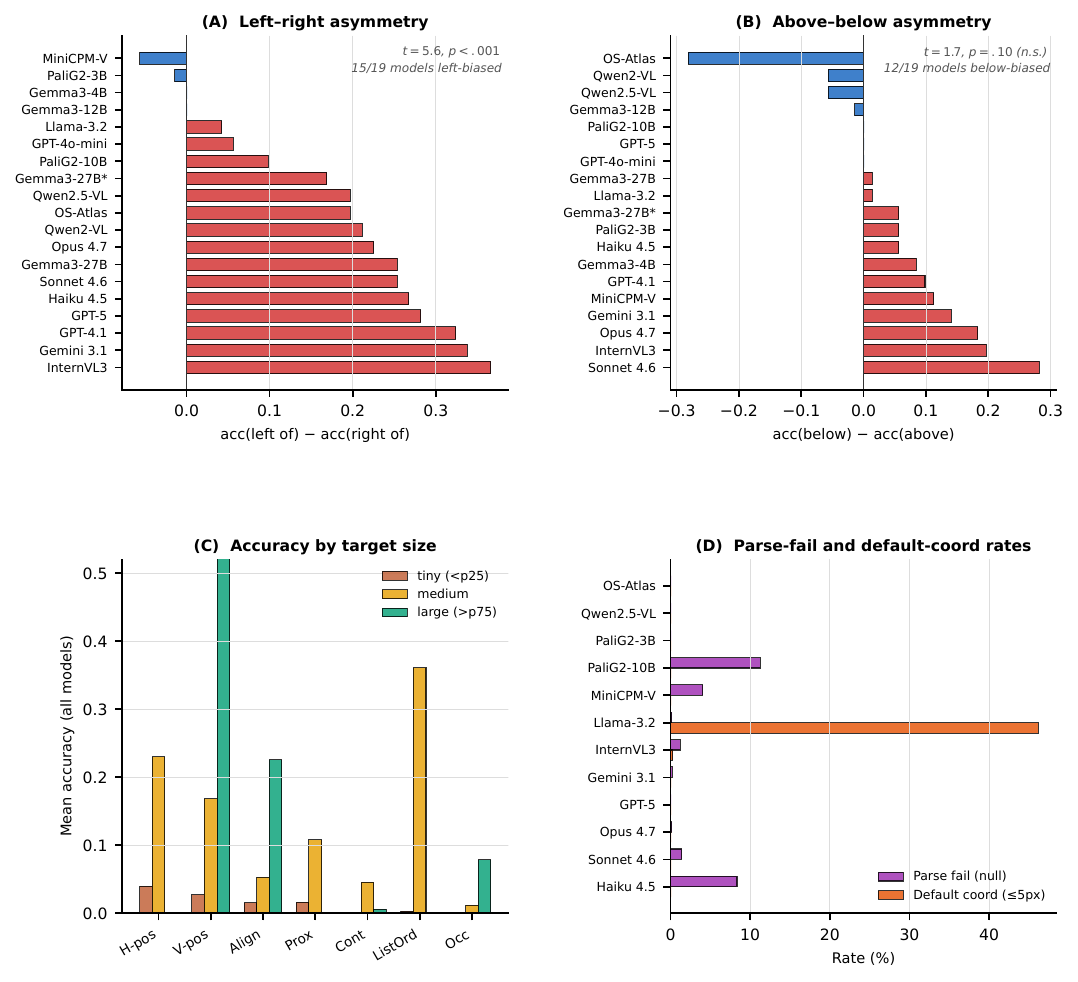}
  \caption{\textbf{Model sensitivity analysis.}
    \textbf{(A)} Left--right accuracy gap per model, sorted by magnitude. Red bars = left-biased (15/19 models; $t = 5.6$, $p < .001$).
    \textbf{(B)} Above--below accuracy gap. Weaker and not significant overall ($p = .10$); OS-Atlas is an outlier with a strong above preference.
    \textbf{(C)} Mean accuracy stratified by target normalized area (tiny/medium/large) per primitive, pooled over all 19 models.
    \textbf{(D)} Parse-fail (null output) and default-coordinate rates per model; both are infrastructure errors, not spatial-reasoning failures.}
  \label{fig:sensitivity}
\end{figure*}

\begin{figure*}[t]
  \centering
  \includegraphics[width=0.8\textwidth]{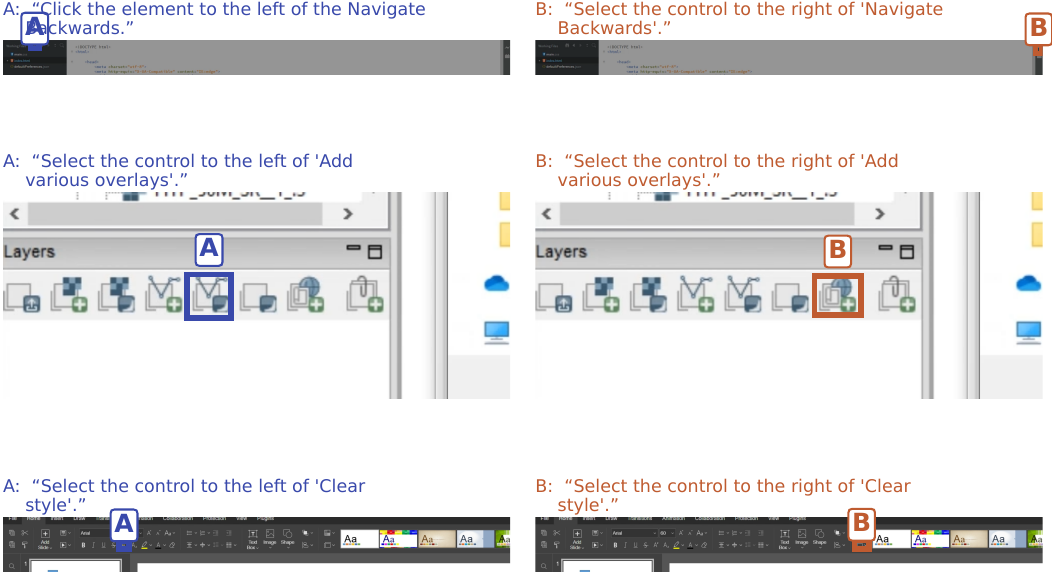}
  \caption{\textbf{\texttt{rel-pos-horizontal} (real screenshots).} Three minimal pairs from UI-Vision. Within each row, both panels show the same screenshot; only the relation word (\emph{right of} vs.\ \emph{left of}), and the gold target moves accordingly. Across rows, the application and anchor change.}
  \label{fig:teaser-rph}
\end{figure*}

\begin{figure*}[t]
  \centering
  \includegraphics[width=0.6\textwidth]{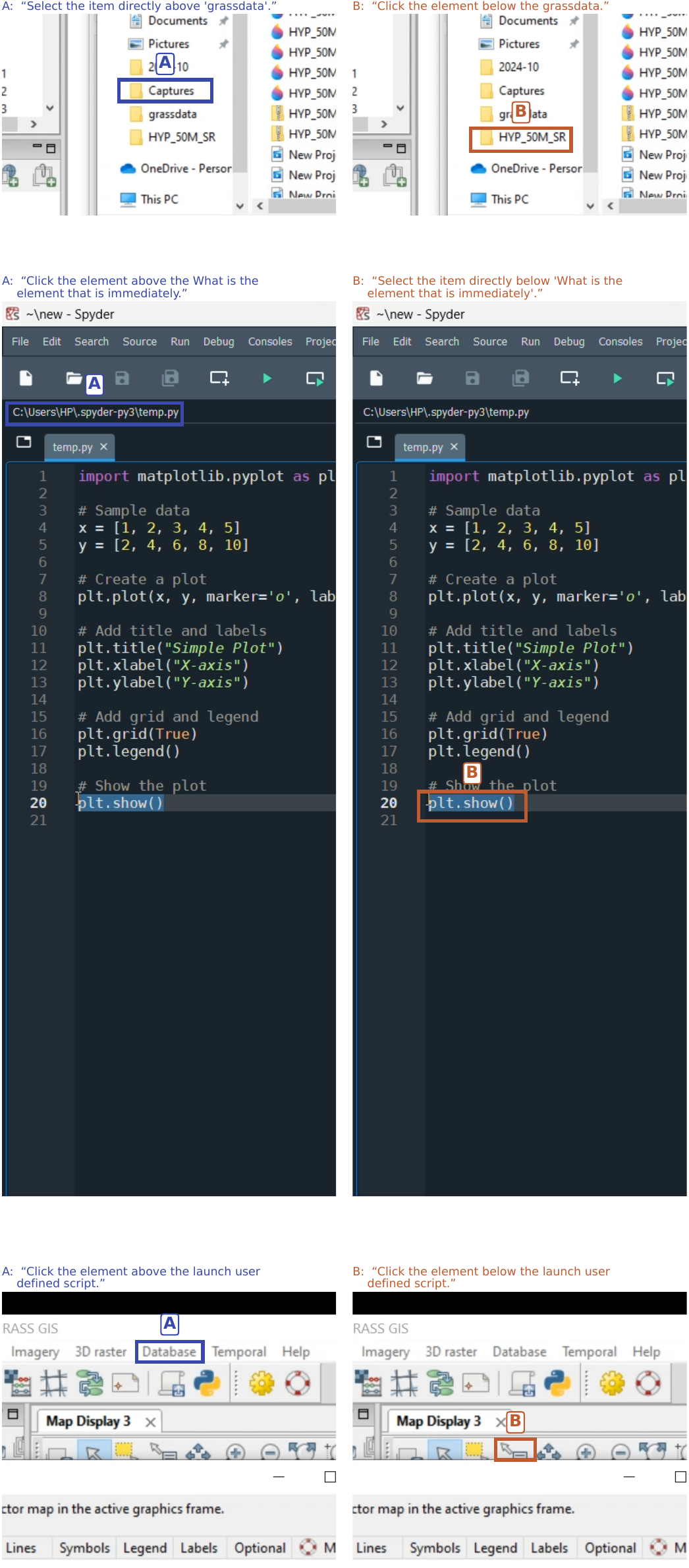}
  \caption{\textbf{\texttt{rel-pos-vertical} (real screenshots).} Three minimal pairs; \emph{above} vs.\ \emph{below} the anchor element across three different real GUIs.}
  \label{fig:teaser-rpv}
\end{figure*}

\begin{figure*}[t]
  \centering
  \includegraphics[width=0.7\textwidth]{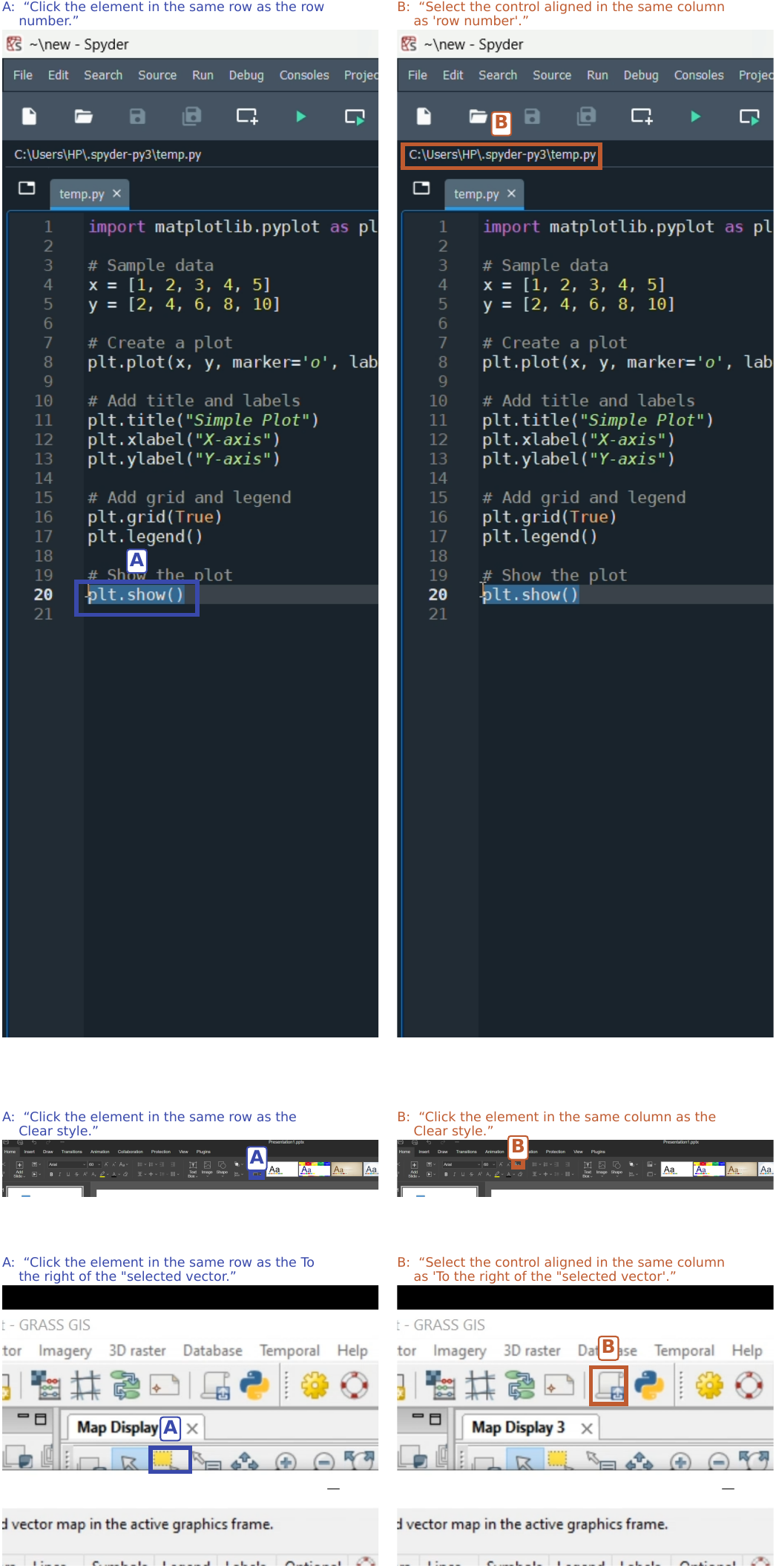}
  \caption{\textbf{\texttt{alignment} (real screenshots).} Three minimal pairs in which the relation switches between \emph{same row as} and \emph{same column as} the anchor.}
  \label{fig:teaser-align}
\end{figure*}

\begin{figure*}[t]
  \centering
  \includegraphics[width=0.96\textwidth]{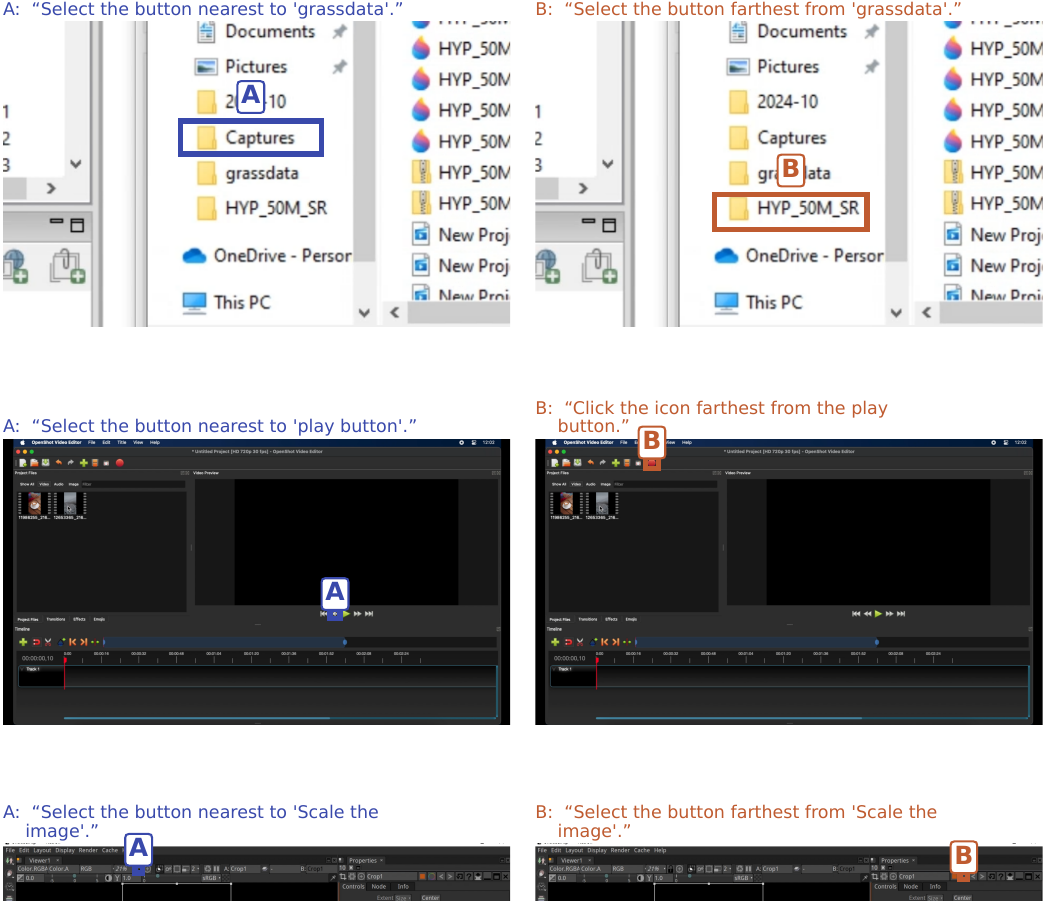}
  \caption{\textbf{\texttt{proximity} (real screenshots).} Three minimal pairs in which the two correct answers are the elements \emph{nearest to} and \emph{farthest from} the same anchor across three different real applications.}
  \label{fig:teaser-prox}
\end{figure*}

\begin{figure*}[t]
  \centering
  \includegraphics[width=0.96\textwidth]{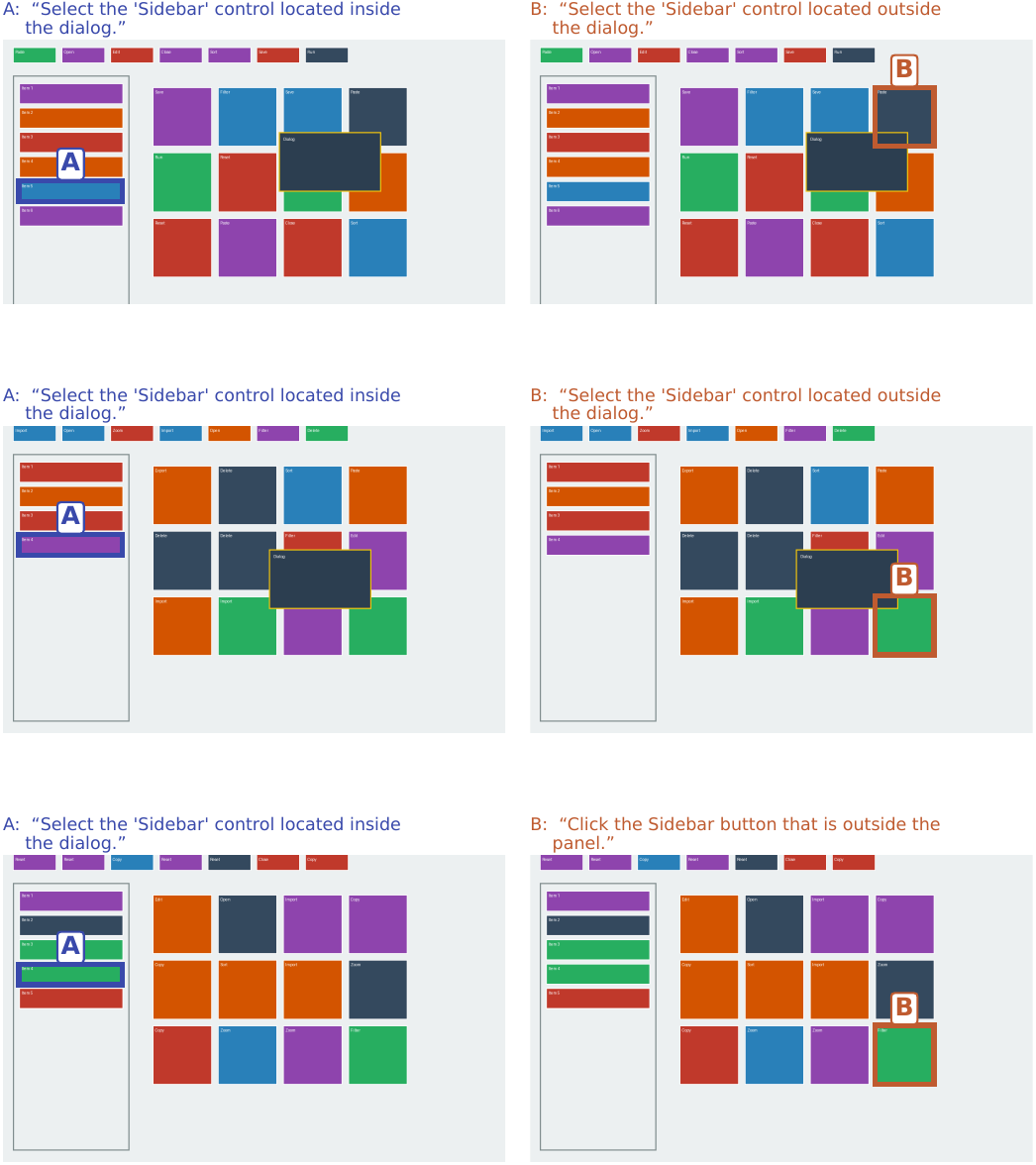}
  \caption{\textbf{\texttt{containment} (synthetic).}
  Three minimal pairs. Twin A is the  \emph{contained within} a parent; twin B is the parent \emph{containing} that inner element. Synthetic stimuli are used because clean parent--child annotations are rare in real screenshot corpora.}
  \label{fig:teaser-cont}
\end{figure*}

\begin{figure*}[t]
  \centering
  \includegraphics[width=0.96\textwidth]{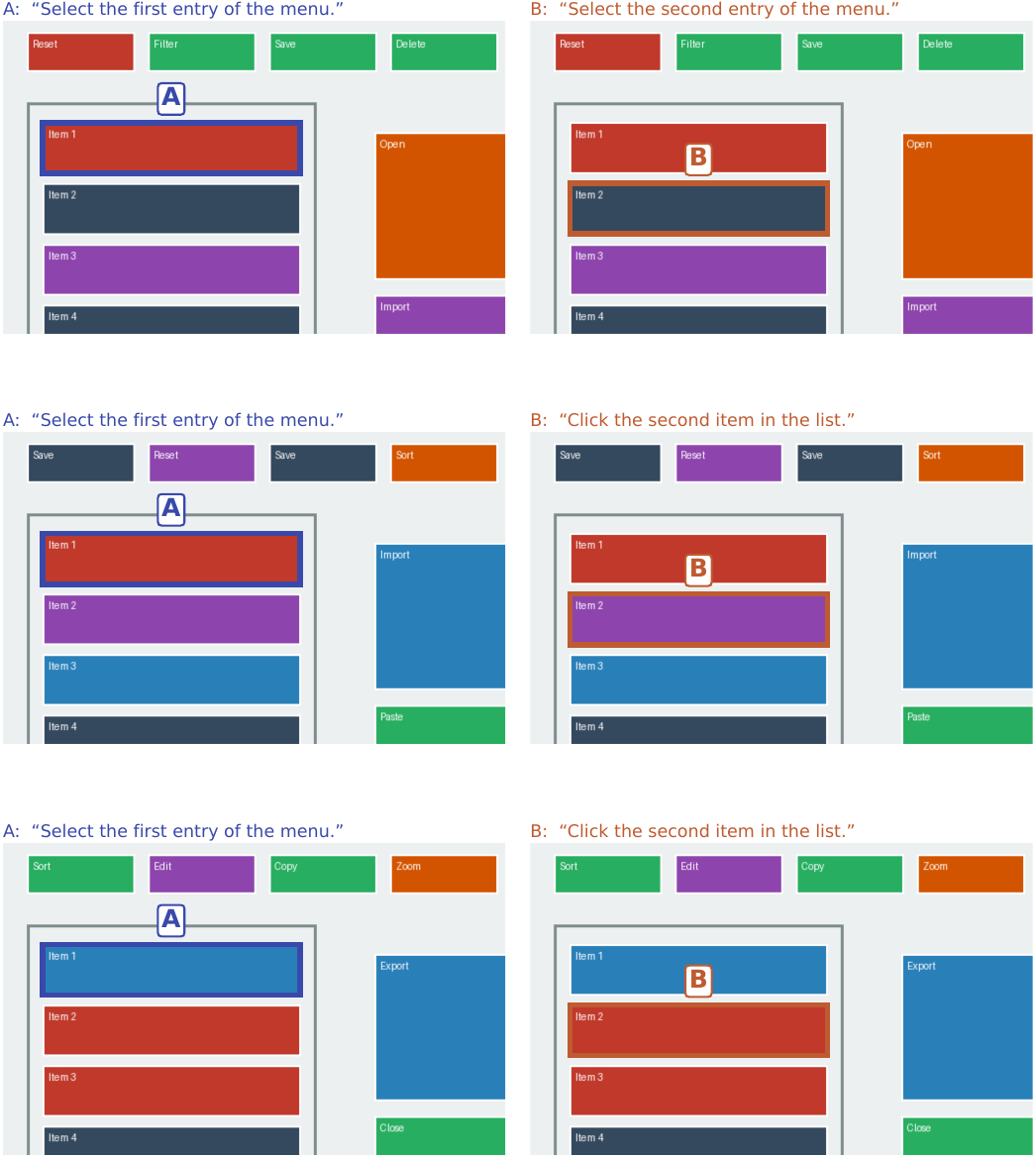}
  \caption{\textbf{\texttt{list-ordinal} (synthetic).} Three minimal pairs of symmetric ordinal queries (the $k$-th item from the top vs.\ from the bottom). The only primitive every top-tier model reliably solves.}
  \label{fig:teaser-list}
\end{figure*}

\begin{figure*}[t]
  \centering
  \includegraphics[width=0.96\textwidth]{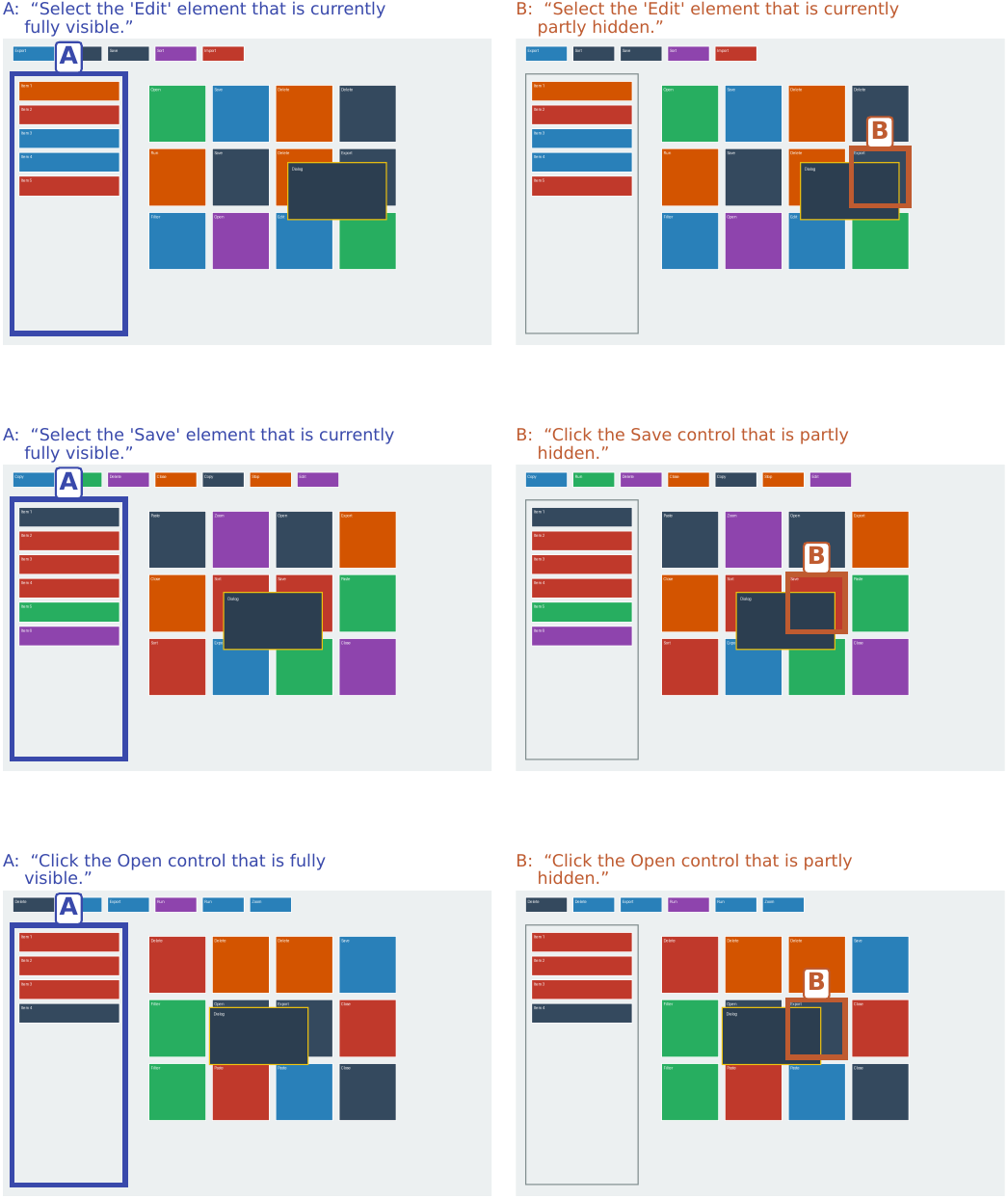}
  \caption{\textbf{\texttt{occlusion} (synthetic).}
  Three minimal pairs. Twin A is the element that is \emph{fully visible}, twin B is the element that is \emph{partly hidden} behind an overlay. A relational-shortcut model is drawn to the visually unusual (occluded) element on both twins, scoring zero at the pair level.}
  \label{fig:teaser-occ}
\end{figure*}

\section{Additional Minimal-Pair Examples}
\label{app:more-teasers}
Figure~\ref{fig:teaser} shows a single \texttt{rel-pos-horizontal} pair. To make the minimal-pair construction intuitive for the other six primitives, Figures \ref{fig:teaser-rph}--\ref{fig:teaser-occ} each present \emph{three} representative pairs for one primitive. Every row is a distinct pair; within a row, both panels share the same screenshot, only the relation word in the instruction differs (twin A on the left, twin B on the right), and the colored box marks only that panel's gold target. A relationally-grounded model must point at the two different boxes for the two instructions; a shortcut-grounded model points at the same box for both.

\begin{figure*}[t]
  \centering
  \includegraphics[width=0.92\textwidth]{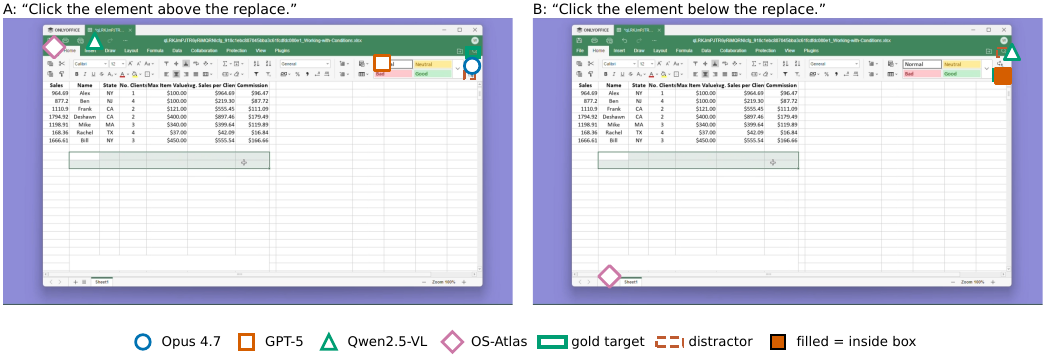}
  \caption{\textbf{Case study 1: \texttt{rel-pos-vertical}, real screenshot.} A: ``Click the element above the replace.'' B: ``Click the element below the replace.'' Opus and GPT-5 collapse to the same coordinate on both twins (both near $(\approx 1180, 130)$), getting B right by luck of which side they defaulted to and missing A. Qwen2.5-VL emits two completely unrelated clicks. Hollow markers = misses; filled markers = hits. Anchor-collapse failure.}
  \label{fig:case01}
\end{figure*}

\begin{figure*}[t]
  \centering
  \includegraphics[width=0.92\textwidth]{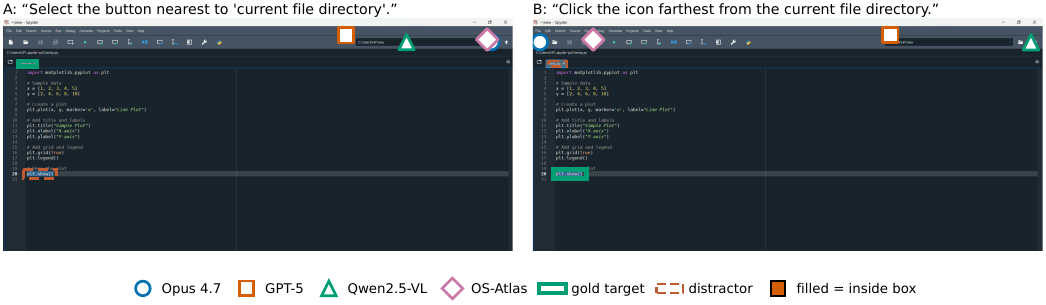}
  \caption{\textbf{Case study 2: \texttt{proximity}, real screenshot.} A: ``Select the button \emph{nearest} to `current file directory'.'' B: ``Click the icon \emph{farthest} from the current file directory.'' All four models cluster their predictions in the top toolbar regardless of the instruction. The anchor (a sidebar element) and the two targets (sidebar items) are far from the predictions. This is the population-level pattern on \texttt{proximity}: 92\% of predictions fall outside both candidate regions (Appendix~\ref{app:candidate}), while the target is selected on 89\% of within-region predictions, locating the deficit in candidate localization rather than relational interpretation.}
  \label{fig:case02}
\end{figure*}

\begin{figure*}[t]
  \centering
  \includegraphics[width=0.92\textwidth]{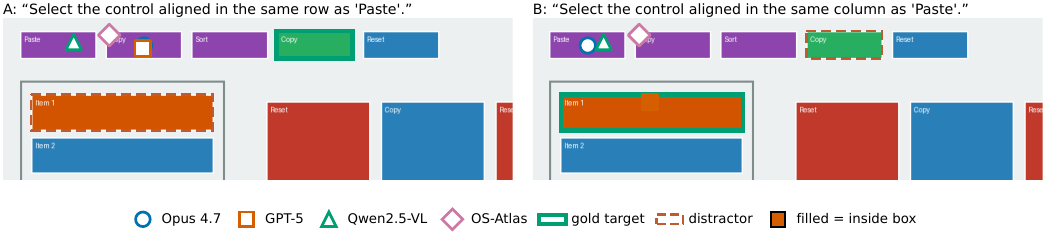}
  \caption{\textbf{Case study 3: \texttt{alignment}, synthetic.} A: ``Select the control aligned in the same \emph{row} as `Paste'.'' B: ``Select the control aligned in the same \emph{column} as `Paste'.'' Models click \emph{the Paste anchor itself} or a visually similar icon, instead of the aligned neighbor. Qwen2.5-VL emits an \emph{identical coordinate} for both twins, the canonical ``relation-ignored'' signature.}
  \label{fig:case03}
\end{figure*}

\begin{figure*}[t]
  \centering
  \includegraphics[width=0.92\textwidth]{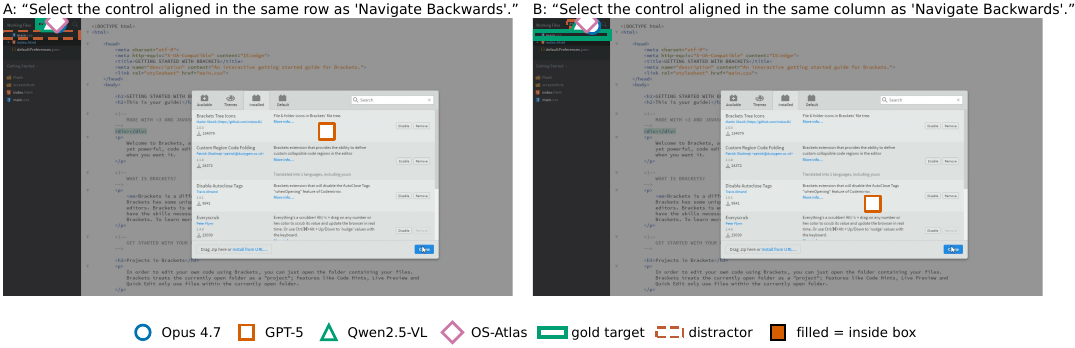}
  \caption{\textbf{Case study 4: \texttt{alignment}, real screenshot.} A/B: ``aligned in the same row/column as `Navigate Backward'.'' The real-screenshot version of case study 3. GPT-5 wanders far from the anchor on both twins. Qwen2.5-VL again emits the same coordinate for both. \texttt{alignment} on real GUIs is the hardest cell in the benchmark.}
  \label{fig:case04}
\end{figure*}

\begin{figure*}[t]
  \centering
  \includegraphics[width=0.92\textwidth]{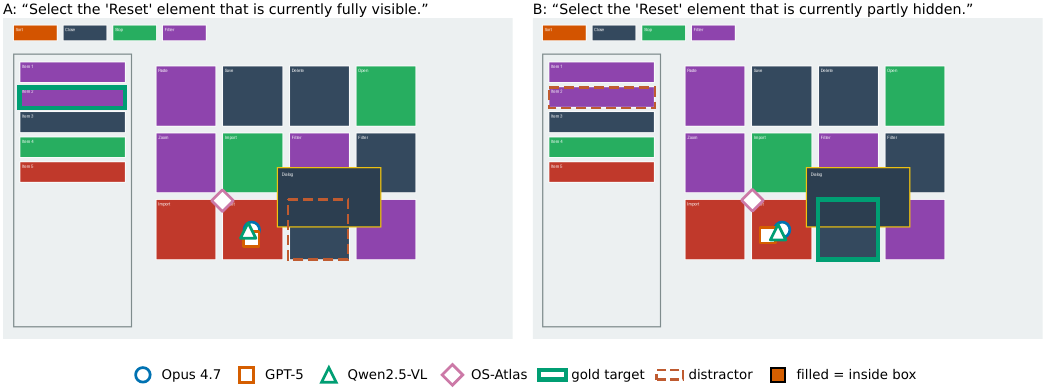}
  \caption{\textbf{Case study 5: \texttt{occlusion}, synthetic.} A: the `Reset' button that is \emph{fully visible}. B: the `Reset' button that is \emph{partly hidden} behind a dialog. All four models point to the partly-hidden button on \emph{both} twins (paired predictions differ by $<$10\,px). The models are drawn to the visually unusual element and never report the visible one. On this pair, the models select the contrastive element on both twins; pooled over all models and items, within-region selection on \texttt{occlusion} is 0.554, statistically indistinguishable from $0.50$ (Appendix~\ref{app:candidate}).}
  \label{fig:case05}
\end{figure*}

\begin{figure*}[t]
  \centering
  \includegraphics[width=0.92\textwidth]{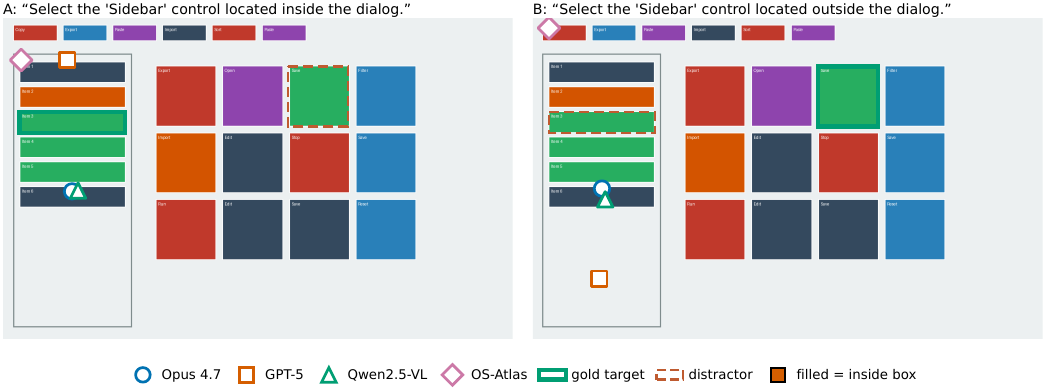}
  \caption{\textbf{Case study 6: \texttt{containment}.} A: target is the element \emph{contained within} the parent. B: target is the parent \emph{containing} the inner element. Models repeatedly click the inner element on both twins, missing the parent on twin B. \texttt{containment} carries the weakest relation-word signal in the benchmark: within-region selection splits 0.548/0.452 between target and distractor, and 92\% of predictions fall outside both candidate regions (Appendix~\ref{app:candidate}).}
  \label{fig:case06}
\end{figure*}

\begin{figure*}[t]
  \centering
  \includegraphics[width=0.92\textwidth]{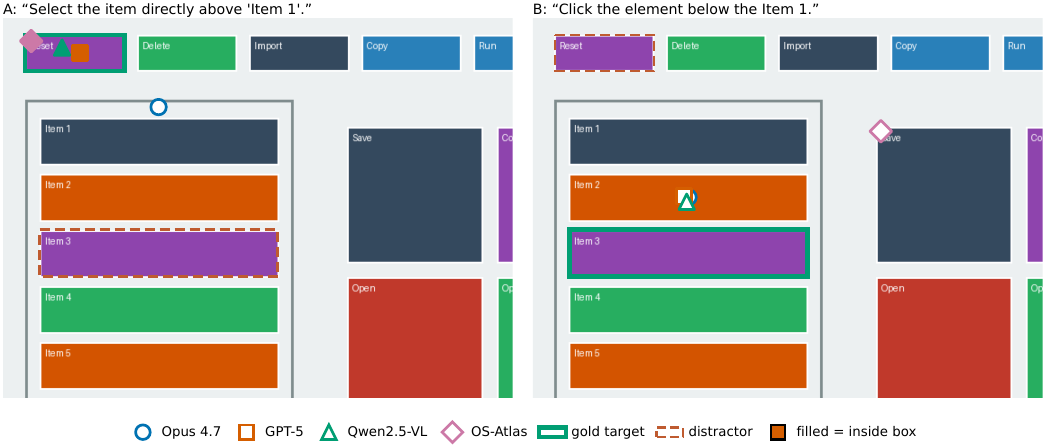}
  \caption{\textbf{Case study 7: \texttt{rel-pos-vertical}, synthetic.} The controlled-stimulus version of case study 1. Even on the clean, synthetic layout, the same anchor-collapse signature is visible: predictions cluster near the anchor instead of resolving above/below.}
  \label{fig:case07}
\end{figure*}

\begin{figure*}[t]
  \centering
  \includegraphics[width=0.92\textwidth]{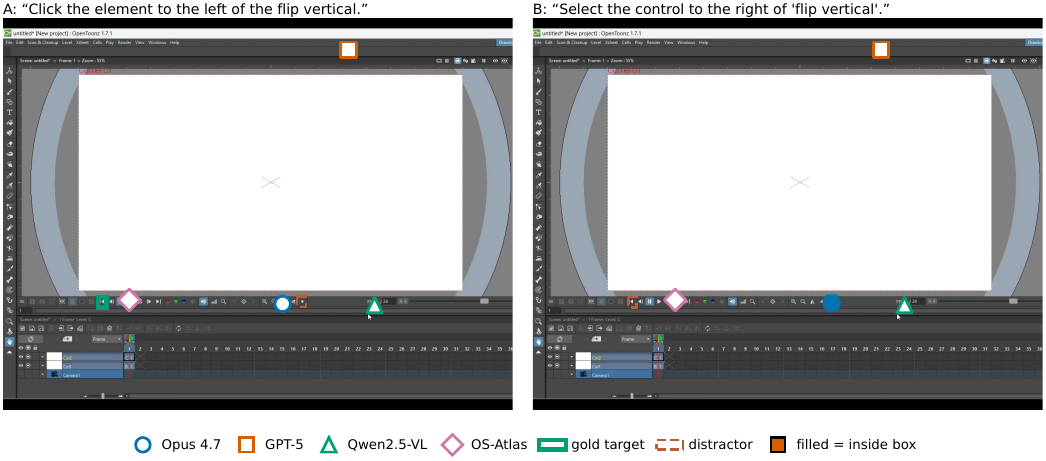}
  \caption{\textbf{Case study 8: \texttt{rel-pos-horizontal}, real screenshot.} A: ``element to the right of \emph{X}''. B: ``element to the left of \emph{X}''. The minimal-pair construction that opens the paper (Fig.~\ref{fig:teaser}) repeats here on a different application. Same shortcut, same failure mode.}
  \label{fig:case08}
\end{figure*}

\section{Model-Specific Sensitivity Analysis}
\label{app:sensitivity}
Figure~\ref{fig:sensitivity} reports four model-level sensitivity diagnostics computed directly from the existing diagnostic runs.

\paragraph{Left--right asymmetry (Panel A).}
Across 19 models, 15 score higher on \texttt{left of} than \texttt{right of} items, with a mean gap of $+0.169$ accuracy points (one-sample $t = 5.6$, $p < .001$, sign test $p < .001$). The bias is present in every proprietary model (Opus, Sonnet, Haiku, GPT-5, Gemini) and in the strongest open models. This is consistent with a left-side layout frequency prior: in Western GUIs, action buttons, primary controls, and frequently clicked elements are disproportionately placed on the left, so models that ground by position salience rather than by relation word will inflate left-side accuracy. The candidate-level analysis (Appendix~\ref{app:candidate}) indicates that this asymmetry coexists with relation-word use rather than replacing it: within-region selection is 0.972 under ``left of'' and 0.717 under ``right of'', and predictions fall within a candidate region 2.7$\times$ more often when the target is the left element. The accuracy asymmetry reported here and the within-region asymmetry in Appendix~\ref{app:candidate} are two measurements of the same directional effect. The two exceptions, MiniCPM-V and PaliGemma2-3B, are low-accuracy models that default to near-origin coordinates on many items (Panel D), effectively averaging away the bias.

\paragraph{Above--below asymmetry (Panel B).}
The below-over-above preference is weaker and not statistically significant (mean $\Delta = +0.047$, $t = 1.7$, $p = .10$, 12/19 models below-biased).
OS-Atlas is a notable outlier with a strong \emph{above} preference ($\Delta = -0.28$), suggesting a GUI-specific training prior that differs from the natural-image convention.

\paragraph{Target-size sensitivity (Panel C).}
Pooled across all models and primitives, large targets ($>$ 75th percentile of normalized area) reach a mean accuracy of 0.092 vs.\ 0.022 for tiny targets ($<$ 25th percentile). The effect is largest for \texttt{containment} and \texttt{list-ordinal}, primitives where the target is a panel or a list row that occupies substantial screen area, and smallest for \texttt{alignment} and \texttt{proximity}, where the spatial relationship matters more than target visibility. Caution: the size gap is partially confounded with source (synthetic targets are larger), so this should not be interpreted as a pure resolution effect.

\paragraph{Coordinate-frame and parse-failure rates (Panel D).}
Three distinct failure modes are visible: Llama-3.2-11B-Vision emits a near-origin default on 46\% of items (consistent with a ``don't know'' fallback, not a parser bug); PaliGemma2-10B and Claude Haiku produce null outputs (empty or unparseable text) on 8--11\% of items; and InternVL3 and Claude Sonnet show low but non-zero null rates. These are infrastructure failures, not spatial-reasoning failures, and are already annotated as Type~C and~F errors in Appendix~\ref{app:errors}.

\section{Candidate-Level Classification of Baseline Predictions}
\label{app:candidate}

The $0.50$ reference level is a two-candidate forced-choice reference rather than the chance level of the prediction task: models emit an unconstrained coordinate, so a prediction can fall outside both candidate regions. To separate selection of the contrastive element from predictions outside both regions, we classify every baseline prediction from all 19 models (18{,}886 records) geometrically: \emph{target} if the point lies inside the target box, \emph{distractor} if inside the distractor box, \emph{neither} if inside neither, and \emph{invalid} if no coordinate could be parsed. A point inside both boxes is assigned to the nearer box center; this case does not occur (0 of 18{,}886). Table~\ref{tab:candidate} reports the four-way split pooled over models, together with the within-region selection rate (target as a fraction of predictions inside either region), its denominator, and a two-sided exact binomial test against $0.50$.

\begin{table}[ht]
\small
\centering
\caption{Candidate-level classification of all baseline predictions, pooled over 19 models (strict rule: a prediction is assigned to a candidate only if it lies inside that candidate's box). Sel.\ = within-region selection rate, target~/~(target~+~distractor), reported with its denominator $n$. $p$ is a two-sided exact binomial test against 0.50. Selection rates computed on a small $n$ (e.g., alignment, $n = 173$) are correspondingly less precise.}
\label{tab:candidate}
\setlength{\tabcolsep}{2.5pt}
\renewcommand{\arraystretch}{1.1}
\resizebox{\columnwidth}{!}{
\begin{tabular}{@{}lrrrrrrl@{}}
\toprule
\textbf{Primitive} & \textbf{Tgt} & \textbf{Dis} & \textbf{Neither} & \textbf{Inv.} & \textbf{Sel.} & $n$ & $p$ \\
\midrule
containment        & 3.2\%  & 2.6\% & 91.9\% & 2.3\% & 0.548 & 157  & 0.26 \\
occlusion          & 6.1\%  & 4.9\% & 88.1\% & 0.8\% & 0.554 & 298  & 0.07 \\
alignment          & 4.7\%  & 1.7\% & 92.5\% & 1.1\% & 0.728 & 173  & $<10^{-8}$ \\
proximity          & 6.1\%  & 0.7\% & 92.4\% & 0.7\% & 0.891 & 184  & $<10^{-28}$ \\
rel-pos-horizontal & 14.8\% & 1.6\% & 83.2\% & 0.4\% & 0.903 & 443  & $<10^{-72}$ \\
rel-pos-vertical   & 16.1\% & 1.7\% & 79.2\% & 3.0\% & 0.904 & 481  & $<10^{-79}$ \\
list-ordinal       & 31.1\% & 6.8\% & 60.3\% & 1.8\% & 0.820 & 1022 & $<10^{-99}$ \\
\bottomrule
\end{tabular}
}
\end{table}

\paragraph{Selection between candidates.}
The target is selected at least as often as the distractor on every primitive. The \emph{neither} column dominates throughout: 60--92\% of predictions fall outside both candidate regions. For \texttt{containment} and \texttt{occlusion}, within-region selection is statistically indistinguishable from $0.50$, indicating no measurable relation-word signal on these two primitives; for \texttt{rel-pos-horizontal}, \texttt{rel-pos-vertical}, and \texttt{proximity}, within-region selection favors the target ($0.89$--$0.90$). At the individual-model level, the only statistically significant excess of distractor selection is Gemma3-4B ($0.328$, $n = 67$); no model shows such an excess under the permissive rule below.

\paragraph{Sensitivity to the inclusion rule.}
The strict rule yields small denominators, so we repeat the classification under a permissive rule: a prediction is assigned to a candidate if it lies within one box-diagonal of that candidate's box center (normalized per box), and to \emph{neither} otherwise. The two rules agree for six of seven primitives: containment 0.539 ($n = 1680$), occlusion 0.557 ($n = 1256$), and the four target-favoring primitives at 0.69--0.77 ($n = 471$--$2075$). The \texttt{alignment} estimate depends on the rule: 0.728 under the strict rule ($n = 173$, 6\% of its predictions) versus 0.529 under the permissive rule ($n = 1021$). We report both values and do not interpret the within-region selection rate for this primitive.

\paragraph{Real-screenshot arm.}
On UI-Vision screenshots, the \emph{neither} column is larger still: pooled over 19 models, 96.2\% of real-arm predictions fall outside both candidate regions (87.8\% for GPT-5 and Claude Opus 4.7 alone), while within-region selection is 0.747 pooled ($n = 150$) and 0.914 for GPT-5 and Opus ($n = 70$). Real-screenshot accuracy is thus primarily attributable to candidate localization rather than to inverted relational interpretation.

\paragraph{Comparison with the human task.}
The human annotation task is a two-candidate forced choice, so the within-region selection rate is the model-side quantity matched to the human accuracy of 96.9\%. On the human-clean core, Claude Opus 4.7 produces a within-region prediction on 62 of 185 items and selects the target on 60 of those 62 (0.968); $60/185 = 0.324$ equals the point-in-box accuracy in Table~\ref{tab:overall}, since ``prediction inside the target box'' is the same event under both definitions. For GPT-5, $54/57 = 0.947$ within-region and $54/185 = 0.292$ overall. Within-region discrimination for the strongest models (0.95--0.97) is comparable to human accuracy (0.969); the difference in overall accuracy is primarily attributable to the low proportion of predictions ($\approx$ one third) that fall within either candidate region.

\paragraph{Directional asymmetry.}
Conditioning within-region predictions on the relation word gives $P(\text{target} \mid \text{``left of''}) = 0.972$ ($n = 323$) versus $0.717$ ($n = 120$) for ``right of''; for the vertical primitive, ``below'' $= 0.984$ ($n = 253$) versus ``above'' $= 0.816$ ($n = 228$). Both conditional rates exceed $0.50$, indicating that the relational word is used on these primitives. The left-side layout prior to Appendix~\ref{app:sensitivity} manifests as a difference in the rate at which predictions fall within a candidate region (2.7$\times$ higher when the target is the left element) together with a lower within-region accuracy under ``right of''.

\paragraph{Cluster-robustness of the below-reference flags.}
The synthetic generator reuses anchor slots (containment: 4 distinct anchors; occlusion: 13; alignment: 34; proximity: 43; 71 screenshots per primitive), so we recompute all per-primitive bootstrap intervals with cluster resampling at the anchor level and at the screenshot level. No upper bound reaches $0.50$ for any of the 19 models on any of the four sub-reference primitives under either clustering; the widest interval is Claude Opus 4.7 on proximity, anchor-clustered, $[0.035, 0.398]$. The below-reference intervals therefore persist under clustering, while the candidate-level classification above indicates that their primary source is off-candidate prediction.

\section{OCR, Layout, and Cross-Modal Binding: Disentangling Failure Sources}
\label{app:binding}

The blur-control experiment (\S\ref{subsec:controls}) shows that removing high-frequency detail, which degrades fine-grained OCR of text labels, reduces accuracy by only 5.3 points, substantially less than removing the screenshot entirely ($-$15.5 points) or shuffling it ($-$7.2 points).
This suggests that models rely more on global layout structure than on reading individual element labels, but it does not isolate \emph{which} component of grounding fails.
Three failure sources are plausible: (1) \textbf{text/OCR dependence}: the model cannot read the anchor label and therefore cannot localise the relational neighbour; (2) \textbf{spatial-encoding failure}: the model reads the anchor correctly but misrepresents its position or the spatial relation in the instruction; (3) \textbf{cross-modal binding failure}: the model reads both the anchor and the relation word correctly but fails to compose them into the correct click target.

\paragraph{What the blur result tells us.}
The modest accuracy drop under blur is evidence against strong OCR dependence for the binary outcome: even when text is illegible, models still achieve 16.7\% (well above the text-only baseline of 6.5\%), implying that global spatial structure alone carries residual information. However, blur does not selectively impair text: it also degrades icon shapes and element boundaries, so we cannot distinguish OCR loss from icon-recognition loss under this control alone.

\paragraph{What targeted ablations would reveal.}
Three ablations would cleanly disentangle the three failure sources, and we lay them out as concrete future directions:
\begin{enumerate}[leftmargin=1.4em, itemsep=2pt]
  \item \textbf{Text masking / selective blur.} Replace all text regions (detected via OCR or the accessibility tree) with uniform grey blocks while preserving spatial layout. A large drop relative to full-image accuracy would indicate OCR dependence; a small drop would point to spatial encoding or binding failures.
  \item \textbf{Layout preservation with degraded text.} Render a version of each screenshot where element shapes and positions are preserved, but all text is replaced with random strings of the same length. If accuracy stays near baseline, the model is primarily using spatial structure; if it drops, text semantics matter.
  \item \textbf{Icon glyph perturbation.} Swap icon images with visually similar but semantically different glyphs while keeping positions and text unchanged. This isolates whether failures arise from icon-identity recognition or from relational reasoning once the elements are identified.
\end{enumerate}
We do not run these ablations here because they require either a high-fidelity text-detection pipeline (for selective masking) or controlled re-rendering of real GUI screenshots (for glyph swapping), neither of which is available in the current environment. We flag them as the highest-priority next steps for understanding whether the bottleneck is perceptual (OCR/icon recognition) or compositional (cross-modal binding of the anchor identity to the spatial-relation word).

\section{Qualitative Case Studies}
\label{app:qualitative}

Figures~\ref{fig:case01}--\ref{fig:case08} present eight concrete minimal-pair failures drawn from the top-4 baseline models. In every figure, the two panels are the two twins of one pair (instruction A on the left, instruction B on the right) sharing the same screenshot. Annotations are programmatic, not hand-drawn:

\begin{itemize}\setlength\itemsep{1pt}
\item \textcolor[HTML]{009E73}{\textbf{green box}}: gold target (from \texttt{items.jsonl}).
\item \textcolor[HTML]{BF5B30}{\textbf{dashed terra box}}: the distractor (the wrong twin of the minimal pair).
\item \textbf{coloured markers}: each model's actual click coordinate from \texttt{\small{runs/diagnostic/<model>/diagnostic.jsonl}}. \textbf{Filled} markers are predictions that fall inside the gold box (a hit); \textbf{hollow} markers are misses.
\end{itemize}

\noindent The case studies span every primitive in both data sources; the chosen pairs are those where at least 5 of 8 model--twin evaluations failed, i.e.\ pairs that the strongest models genuinely struggle with. Recurring failure modes are summarized at the end.

\paragraph{Cross-case summary.}
Three patterns recur across all eight case studies and motivate the error analysis in Appendix~\ref{app:errors}: \textbf{(i) Anchor-collapse}: predictions cluster on the anchor noun rather than on its spatial neighbor (cases 1, 3, 4, 7). \textbf{(ii) Twin-identical predictions}: the two paired clicks land within a few pixels of each other, meaning the relation word was ignored (cases 3, 4, 5). \textbf{(iii) Off-candidate prediction}: the dominant population-level failure is that predictions fall outside both members of the pair (60--92\% of predictions; Appendix~\ref{app:candidate}). Case 5 illustrates selection of the contrastive twin at the item level; pooled across all models and items, within-region selection is near $0.50$ on \texttt{containment} and \texttt{occlusion} and favors the target on the remaining primitives.

\begin{figure*}[t]
  \centering
  \includegraphics[width=0.96\textwidth]{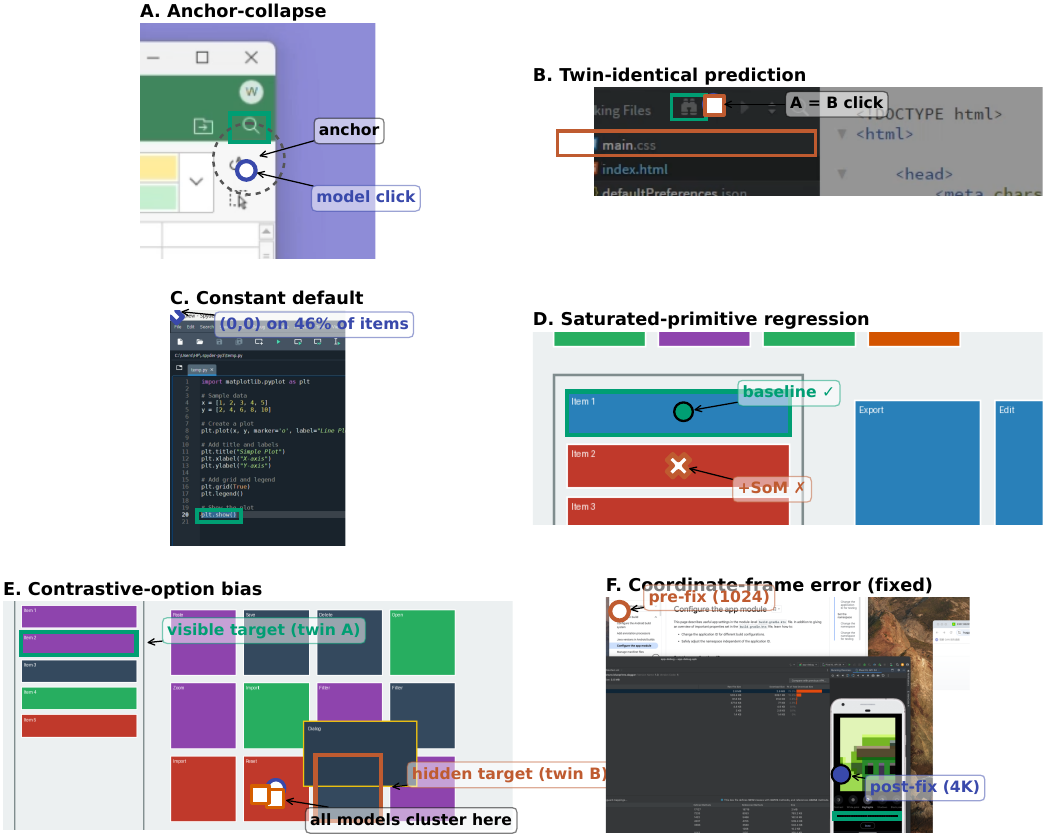}
  \caption{\textbf{Error-type cheatsheet: one real example per error type, drawn from the actual prediction JSONLs.} Right-of-marker labels sit in 50\%-transparent boxes so the underlying screenshot remains visible. \textbf{(A) Anchor-collapse}: \emph{click lands on the anchor noun, not the relational neighbor.} Opus~4.7's prediction on \texttt{ui\_vision-199-0} (``click the element above the replace'') falls inside the dashed ``anchor'' zone (the ``replace'' button), not the green target above it. \textbf{(B) Twin-identical prediction}: \emph{same click for both twins → relation word ignored.} Qwen2.5-VL emits the same click on the two alignment twins of pair 81; both targets are shown for context. \textbf{(C) Constant default}: \emph{model defaults to a canonical coordinate.} Llama-3.2-V emits literal $(0,0)$ on $\sim$46\% of diagnostic items; the prediction marker sits in the upper-left corner regardless of where the target is. \textbf{(D) Saturated-primitive regression}: \emph{intervention hurts a primitive the model already solved.} On \texttt{list-ordinal} the green-filled marker shows that Qwen2.5-VL's baseline answer was correct, but the SoM adds an indirection and the same model misses (red ${\times}$). \textbf{(E) Contrastive-option choice}: \emph{on this pair, the models select the contrastive target on both twins.} On the \texttt{occlusion} pair, both Opus (circles) and GPT-5 (squares) cluster on the partly-hidden target on \emph{both} twins, ignoring the visible target. Pooled over all models and items, within-region selection on \texttt{occlusion} is statistically indistinguishable from $0.50$ (Appendix~\ref{app:candidate}). \textbf{(F) Coordinate-frame error (now fixed)}: \emph{API returns coords in a downsampled frame.} GPT-5's pre-fix ScreenSpot-Pro prediction (terra hollow circle) lands in a 1024-side frame far from the 4K target; after our pre-resize + scale-back wrapper, the post-fix prediction (indigo filled circle) sits near the gold box.}
  \label{fig:err-cheatsheet}
\end{figure*}

\section{Error Analysis}
\label{app:errors}
We classify every failed prediction (across the 19~$\times$~994 $=$ 18{,}886 model--item evaluations on the full benchmark) into one of six error types. Figure~\ref{fig:err-cheatsheet} shows one real example of each type drawn from the model predictions. The types are not mutually exclusive; a prediction can belong to more than one, but they are intended to be informative.

\paragraph{Type A: Anchor-collapse.}
The prediction lands within $1{\times}$ the diagonal of the anchor element's bounding box and not in the target box. The model correctly locates the anchor (the noun the instruction refers to), but emits a click on the anchor itself rather than on the spatial neighbor the relation word specifies. Most common on \texttt{rel-pos-horizontal} and \texttt{rel-pos-vertical} (Example 1 above). Affects 28\% of all failed binary-primitive items on Opus 4.7, 21\% on GPT-5.

\paragraph{Type B: Twin-identical prediction.}
The two paired items receive predictions whose Euclidean distance in pixel space is less than 10\% of the smaller side of the screenshot. The model emitted essentially the same click for ``left of X'' and ``right of X'', telling us that the relation word was ignored. This is the most diagnostic failure mode: it pushes \emph{pair-consistency} far below \emph{accuracy} (the ``\textsc{What's-Up} signature'' of \citealp{kamath2023whatsup}, re-derived in \S\ref{subsec:pair-consistency} / Fig.~\ref{fig:pair}). Affects 37\% of paired failures on Qwen2.5-VL, 25\% on Opus 4.7, and 41\% on GPT-5 \emph{before} the closed-API coordinate rescale; after the rescale (Appendix \ref{app:rescale}), the GPT-5 rate drops to 24\%.

\paragraph{Type C: Constant default.}
The model emits a fixed canonical coordinate (e.g.\ $(0,\,0)$ or the image center) on a large fraction of items. The clearest case is Llama-3.2-11B-Vision, which emits $(0,\,0)$ on 46\% of diagnostic items and 25\% of ScreenSpot-Pro items. This is a model-side ``don't-know'' fallback, not a parser failure: \texttt{raw\_text} is a well-formed \texttt{click(0,\,0)} string. We document it because the headline 1.2\% accuracy of Llama-3.2 on \bench{} \emph{understates} the model: its actual decision is ``refuse to ground'' on roughly half the items.

\paragraph{Type D: Saturated-primitive regression.}
The model already solves the primitive without the intervention, so the intervention \emph{introduces friction}. The clearest instance is SoM on Qwen2.5-VL's \texttt{list-ordinal}, which regresses by 32.4 points (\S\ref{sec:interventions}, Appendix \ref{app:som-failures}). The intervention converts a task the model had already learned (counting list items) into a mark-selection task with one indirection. The lesson is that training-free interventions should be applied selectively rather than uniformly.

\paragraph{Type E: Contrastive-option choice (item level).}
On individual pairs, several models select the contrastive element on both twins (Example 5). At the population level, the candidate-level classification in Appendix~\ref{app:candidate} shows that below-reference point-in-box accuracy is primarily attributable to predictions outside both candidate regions: pooled over 19 models, the target is selected at least as often as the distractor on every primitive. For \texttt{containment} and \texttt{occlusion}, within-region selection is 0.548 and 0.554, statistically indistinguishable from $0.50$, indicating no measurable relation-word signal on these two primitives. The directional layout prior is quantified in Appendix~\ref{app:sensitivity}: predictions concentrate on left-side elements, and within-region accuracy is lower under ``right of'' than ``left of''.

\paragraph{Type F: Coordinate-frame error (now fixed).}
Closed-API providers downsample large screenshots and return coordinates in the downsampled frame. Before our wrapper handled this, predictions on 4K ScreenSpot-Pro images landed in a $\sim$1024-side coordinate space and missed every target. This was the dominant error category for Claude Sonnet 4.6 (1{,}297\,/\,1{,}581 SS-Pro records) and for the first GPT-5 ScreenSpot-Pro run ($\approx$1{,}500\,/\,1{,}581 records); both are corrected in the version reported in this paper (Appendix \ref{app:rescale}). We flag this category here because future closed-VLM evaluations that omit the rescale will misattribute this error to genuine model failure.

\paragraph{Cross-cutting observation.}
Types A, B, and E share a common cause: the model has correctly \emph{detected} the elements but failed to \emph{compose} them with the relation word. Types C and F are infrastructure/fallback errors. Type D is an intervention-design issue. The fact that training-free interventions move the needle only on the composition-side errors, and only when the intervention bypasses the relation (SoM) rather than scaffolding it (CoT, steering), is consistent with the \emph{discussion} section's reading that the bottleneck is on the perception / binding side, not on the language side.

\section{Limitations}
\label{app:limits}

\paragraph{Two primitives are synthetic-only.}
Containment and occlusion are present only in the synthetic arm of \bench{}; list-ordinal draws 24 of its 142 items from real UI-Vision menus and lists, with the remainder synthetic. Real GUI datasets do not provide scalable parent--child or overlay--target metadata, so containment and occlusion cannot be mined from real screenshots. Consequently, findings on these two primitives have limited external validity beyond settings where controlled synthetic stimuli are representative of real GUI behavior; the failure rates observed on containment and occlusion may not transfer directly to uncurated real-world GUIs. This constraint bounds the candidate-level result of Appendix~\ref{app:candidate}: the absence of a measurable relation-word signal on containment and occlusion is established on synthetic screenshots only. We treat the synthetic items as a controlled-stimulus arm in the \textsc{What's-Up} tradition and report all results split by source arm.

\paragraph{Per-primitive regression is conservative.}
After the model fixed effects absorb between-model variance, individual per-primitive coefficients in our item-level regression are small. We therefore claim only the joint, model-level association and do not interpret per-primitive coefficients causally. The model-level Spearman is statistically significant; the per-primitive forest plot is reported for transparency, not as a contribution.

\paragraph{No OSWorld downstream slice.}
We stop at ScreenSpot-Pro. End-to-end task success on OSWorld \citep{xie2024osworld} and its verified release \citep{osworldverified2025} would provide stronger downstream confirmation, but it is a separate VM-harness study and is left for future work.

\paragraph{Action and language scope.}
\bench{} v1 covers English instructions and click targets. Future versions could extend the diagnostic to drag-target prediction, bidirectional and CJK layouts, and an OSWorld pre-action grounding slice.

\paragraph{SoM is a scaffold, not a fix.}
SoM partially bypasses the underlying perception-and-binding problem by converting relational grounding into symbolic mark selection; the observed gains reflect scaffolding rather than repair of the spatial skill gap. Additionally, our setup overlays at most two candidate marks per item (target and distractor), which is far fewer than the crowded mark sets that arise in realistic GUI scenes. Gains observed under this minimal two-mark setup may therefore not generalize to settings with many candidates, where disambiguating among marks becomes a bottleneck in its own right. Finally, in deployment, marks would be proposed by an OmniParser-style or accessibility-tree-driven component, with its failure modes propagating upstream.

\paragraph{Point-in-box accuracy is a coarse metric.}
Point-in-box accuracy, while standard for GUI grounding, is binary: a click one pixel outside the target box and a click on the entirely wrong element score identically. The loose metric (within $2\times$ target diagonal of the box center) partially separates sub-pixel imprecision from element misidentification, but neither metric captures the structure of spatial errors: whether the model was on the right side of the screen, how far it drifted, or whether it collapsed to the anchor rather than the relational neighbor. These structured error patterns are documented qualitatively in Appendix~\ref{app:errors}, but a finer-grained spatial-error metric remains future work.

\paragraph{Llama-3.2-Vision returns default coordinates.}
Llama-3.2-11B-Vision emits a constant or near-origin coordinate on roughly 42\% of items. The 1.2\% accuracy is the model defaulting, not a parsing bug; it is consistent with the original ScreenSpot-Pro
paper's report of $<\!2$\% accuracy for generalist VLMs.

\paragraph{Proprietary API behavior may shift.}
The proprietary providers can change downsampling, decoding, or model identity at any time. We freeze a snapshot of dates and pin model identifiers in Appendix~\ref{app:models}; future runs may differ.

\section*{Information About Use Of AI Assistants}
AI assistants were used for limited editing support and language refinement. The final manuscript contents were verified and authored by the research team.

\end{document}